\documentclass[letterpaper]{article}
\usepackage{aaai2027} 
\nocopyright 
\usepackage[hyphens]{url}
\usepackage{graphicx}
\graphicspath{{figures/}}
\usepackage{natbib}
\usepackage{caption}
\usepackage{algorithm}
\usepackage{algorithmic}
\usepackage{multirow}
\usepackage{amsmath}
\usepackage{amssymb}
\usepackage{pifont}
\usepackage{booktabs}
\usepackage[table]{xcolor}
\usepackage{subcaption}
\usepackage{enumitem}
\usepackage{tikz}
\usepackage{microtype}
\usepackage{bbold}

\usetikzlibrary{arrows.meta, positioning, calc, shapes.geometric, fit, backgrounds, decorations.pathreplacing}

\title{SyncPlan: Long-Horizon LLM Coordination with Explicit Synchronization and Adaptive Correction}

\author{
Shen You\textsuperscript{\rm 1,*},
Xiaoming Zhu\textsuperscript{\rm 2,*},
Weining Weng\textsuperscript{\rm 2},
Hefei Mei\textsuperscript{\rm 1},
Weixuan Wang\textsuperscript{\rm 2},
Zhongshen Li\textsuperscript{\rm 1},
Zeji Li\textsuperscript{\rm 1}, \\
Ye-Wen Wang\textsuperscript{\rm 2},
Zijun Liao\textsuperscript{\rm 2},
Juchao Zhuo\textsuperscript{\rm 2},
Yang Wei\textsuperscript{\rm 2},
Fuhao Qiu\textsuperscript{\rm 2},
Siqin Li\textsuperscript{\rm 2},
Zhenjie Lian\textsuperscript{\rm 2}, \\
Danei Gong\textsuperscript{\rm 1},
Junkai Ji\textsuperscript{\rm 3},
Xiangtao Li\textsuperscript{\rm 4},
Qiuzhen Lin\textsuperscript{\rm 3},
Liang Wang\textsuperscript{\rm 2,**},
Ka-Chun Wong\textsuperscript{\rm 1,**}
}
\affiliations{
\textsuperscript{\rm 1}City University of Hong Kong,
\textsuperscript{\rm 2}Tencent,
\textsuperscript{\rm 3}Shenzhen University,
\textsuperscript{\rm 4}Jilin University,

}

\begin{document}

\maketitle

\begin{abstract}

LLM-based multi-agent coordination faces a fundamental trade-off between efficiency and adaptivity in dynamic environments. Existing approaches typically rely on repeated LLM invocations or multi-round communication to adapt decisions during execution, introducing substantial latency and making coordination vulnerable to asynchronous progress and environmental changes. Conversely, one-shot planning reduces coordination overhead but produces open-loop plans that can quickly become stale or fail when actions depend on other agents and the environment.
We introduce \emph{\textbf{SyncPlan}}, a plan-execute-correct framework for long-horizon coordination through explicit synchronization and adaptive correction. Given the state and team-level task, a centralized LLM coordinator generates per-agent action chains in a single planning call. During execution, explicit wait primitives and deadlock detection enforce inter-agent and agent-environment dependencies, while a lightweight Plan Staleness Detector continuously assesses the remaining plan and triggers replanning when environmental changes invalidate its assumptions. We further optimize the coordinator through SFT and planning-oriented RL with dense task progress and outcome-level execution feedback.
Experiments on the public Overcooked benchmark and the complex Honor of Kings environment show that SyncPlan achieves state-of-the-art task success rates while using less than $0.05\%$ of the wall-clock runtime compared with existing LLM-based coordinators.
Code and datasets will be made publicly available.

\end{abstract}

\section{Introduction}
Large Language Models (LLMs) provide a flexible way to coordinate multiple agents toward shared objectives in game environments~\cite{tig2025,shridhar2023distilling,sima2_2025}.
We study a centralized setting in which an LLM coordinator observes the environment state and generates high-level action chains for multiple low-level agents when given a team-level task.
Example tasks include eliminating a designated enemy in a Multiplayer Online Battle Arena (MOBA) game or coordinating two agents to prepare and deliver a dish in Overcooked.
These environments evolve continuously due to stochastic events, actions of non-controlled agents, and state changes of task-relevant entities, requiring adaptive coordination throughout execution.

In this setting, Traditional Multi-Agent Reinforcement Learning (MARL) methods~\cite{rashid2018qmix,yu2022mappo,kuba2022happo} require extensive task-specific interaction and careful reward engineering, while complex coordination preferences are difficult to encode as scalar rewards.
Recent LLM-based coordinators~\cite{capo2025,proagent2024} offer a more expressive alternative by interpreting task objectives and generating explicit team-level plans.
However, they typically rely on repeated LLM invocations~\cite{proagent2024} or multi-round communication~\cite{capo2025} during execution, introducing latency and cost that are acceptable in low-tempo settings but impractical for real-time multi-agent coordination.

\begin{figure}[t]
 \centering
 \includegraphics[width=\linewidth]{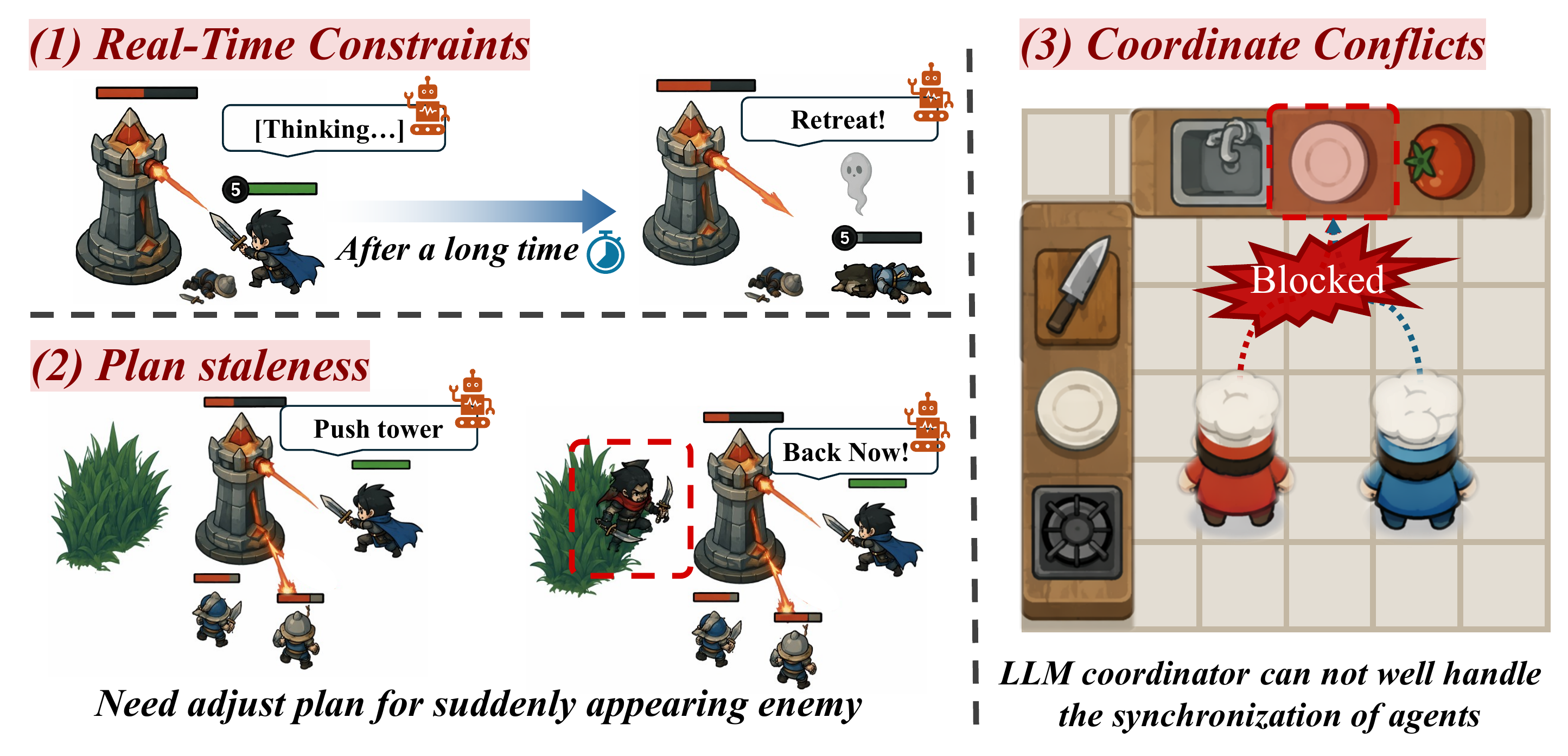}
 \caption{\textbf{Core challenges} of multi-agent coordination: real-time constraints, plan staleness, and coordination conflicts. }
 \label{fig:pec_motivation}
\end{figure}

As illustrated in Fig.~\ref{fig:pec_motivation}, this setting introduces three main challenges:
\begin{itemize}
\item \textbf{Real-time Constraints.}
Real-time environments evolve at the frame level, while frequently invoking LLM coordination incurs intolerable computational overhead under limited response-time budgets.

\item \textbf{Plan staleness.}
Environment changes may invalidate a joint plan. Delayed replanning causes agents to continue executing stale actions, whereas frequent or periodic replanning wastes computation when the plan remains valid.

\item \textbf{Coordination conflicts.}
Asynchronous execution introduces temporal and conditional dependencies, such as waiting for teammates or environmental conditions. Yet LLM plans express them in free-form text that low-level executors cannot reliably verify or enforce.
\end{itemize}
Therefore, the core issue is: \textbf{How can LLM-based multi-agent coordination remain efficient, synchronized, and adaptive during execution?}

To address these challenges, we propose \textbf{SyncPlan} (\textit{\textbf{Sync}hronized \textbf{Plan}ning}), a plan-execute-correct framework for long-horizon multi-agent coordination. In one-time LLM calling, the coordinator produces a long-horizon joint plan comprising structured action chains that span multiple environment frames, amortizing planning cost over extended execution.

A runtime executor advances these chains asynchronously and enforces inter-agent and agent-environment dependencies through executable synchronization primitives. These primitives make temporal dependencies explicit: an agent proceeds only after the required teammate milestone or environment condition is satisfied. The executor further monitors wait dependencies among agents and triggers replanning when cyclic dependencies cause execution to deadlock.

To maintain plan validity as the environment evolves, SyncPlan introduces a lightweight Plan Staleness Detector (PSD) that evaluates the remaining plan against current entity states. It invokes the coordinator when the current plan becomes invalid, thereby retaining adaptivity without periodic LLM inference.

We further optimize the coordinator through SFT and planning-oriented RL. SFT warm-starts the coordinator with structured multi-agent planning capabilities, while RL optimizes complete action chains using dense task progress and outcome-level execution feedback, alleviating long-horizon credit assignment and stabilizing policy optimization.

On the public Overcooked benchmark, SyncPlan improves task achievement rate by an average of \textbf{12.2 percentage points} over the strongest LLM-based baseline in each setting, while requiring less than
\textbf{0.05\%} of their wall-clock runtime. On the substantially more complex Honor of Kings testbed, SyncPlan achieves \textbf{86.3\%} task achievement, surpassing DPT-Agent by \textbf{17.6 percentage points} and reducing relative runtime by over \textbf{26$\times$}, enabling real-time coordination in a 5v5 MOBA environment.

In summary, our main contributions are the following:
\begin{enumerate}
 \item We introduce executable synchronization primitives that convert implicit coordination intent into explicit constraints for asynchronous execution.

 \item We propose a lightweight PSD model that detects when the remaining plan becomes invalid and triggers on-demand replanning.

 \item We propose progressive composite rewards that jointly optimize task effectiveness, execution correctness, and runtime efficiency, providing dense intermediate feedback to improve long-horizon reinforcement learning.

 \item Experiments on Overcooked and Honor of Kings demonstrate state-of-the-art task achievement with substantially lower runtime, enabling real-time 5v5 coordination.

\end{enumerate}

\section{Related Work}

\paragraph{Multi-Agent Reinforcement Learning.}
MARL achieves coordination through joint policy optimization under CTDE, including value decomposition (QMIX~\cite{rashid2018qmix}, QTRAN~\cite{son2019qtran}) and policy-gradient methods (MAPPO~\cite{yu2022mappo}, HAPPO~\cite{kuba2022happo}). Hierarchical variants such as HAVEN~\cite{haven2023} and RHMC~\cite{rhmc2024} introduce sub-goal or sub-action abstraction.

\paragraph{LLM-Based Planning and Multi-Agent Coordination.}
Single-agent LLM planning has been studied in robotics (SayCan~\cite{saycan2022}, Inner Monologue~\cite{innermonologue2022}), open-world games (Voyager~\cite{voyager2023}, SIMA\,2~\cite{sima2_2025}), and hero-level decision-making (TiG~\cite{tig2025}, SIMA~\cite{sima2_2025}). In particular, ReCAPA~\cite{recapa2025} further performs hierarchical predictive correction through repeated LLM invocation, but focuses on a single agent and does not address concurrent multi-agent synchronization.
For multi-Agent coordination, prior work spans dialogue-based consensus (CaPo \cite{capo2025}, MAGRPO~\cite{magrpo2026}), hierarchical plan-then-execute pipelines (MindAgent~\cite{mindagent2023}, L2M2~\cite{l2m2_2025}), world-model-based cooperation (COMBO~\cite{combo2025}), and teammate modeling (ProAgent~\cite{proagent2024}). Their coordination dependencies are generally implicit, while execution-time adaptation relies on repeated LLM invocation or lacks an explicit signal for when the joint plan should be revised.

\begin{figure*}[!htbp]
\centering
\includegraphics[width=0.98\textwidth]{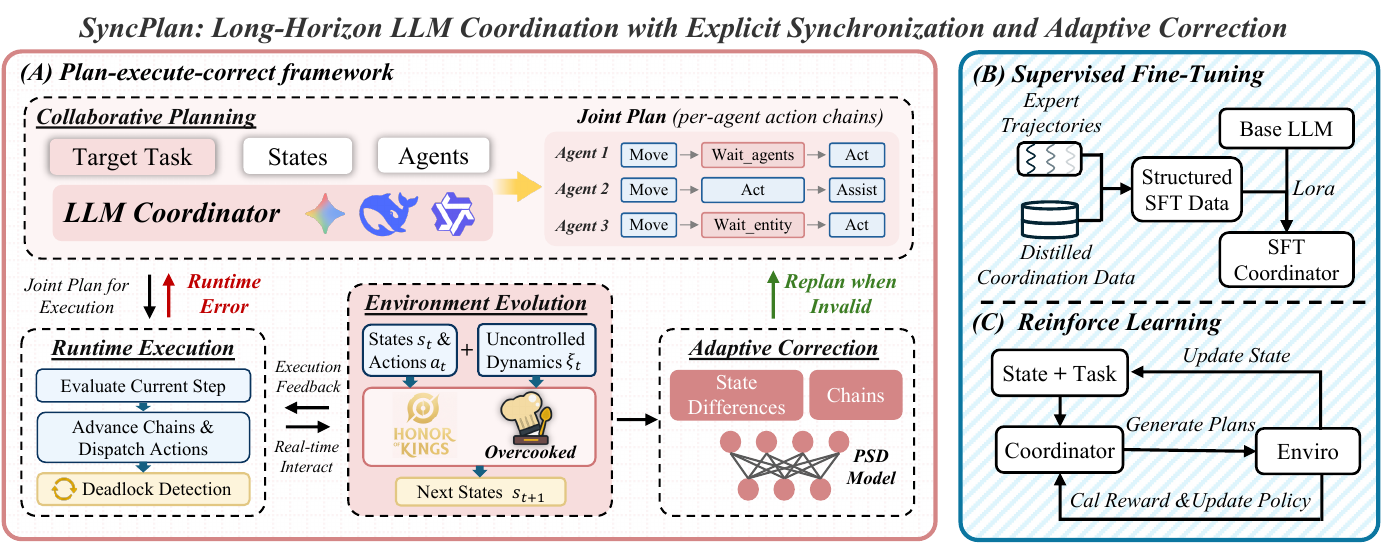}
\caption{\textbf{Overall framework}. \textbf{(A)} SyncPlan forms a plan-execute-correct loop: collaborative planning generates per-agent action chains; runtime execution advances the chains and detects deadlocks; and the PSD monitors plan validity. \textbf{(B--C)} SFT warm-starts the coordinator with expert and distilled data, followed by planning-oriented RL with execution feedback.}

\label{fig:pec_overview}
\end{figure*}

\section{Problem Formulation}
\label{sec:problem}
We consider a dynamic environment with $N$ controlled low-level agents. At frame $t$, the environment state is in state $s_t\in\mathcal{S}$ and evolves according to the transition function
\begin{equation}
 s_{t+1}=\mathcal{T}(s_t,\mathbf{a}_t,\xi_t)
\end{equation}
where $\mathcal{T}:\mathcal{S}\times\mathcal{A}^N\times\Xi\to\mathcal{S}$ is the state transition function and $\mathbf{a}_t=\{a_t^i\}_{i=1}^{N}$ denotes the joint low-level action of the controlled agents. And $\xi_t\in\Xi$ represents uncontrolled stochastic dynamics, such as actions of non-controlled agents and unexpected environment events, and $\xi_t\in\Xi$ is highly related to the state-dependent distribution $\xi_t\sim p(\cdot|s_{t})$.

Each episode is associated with a team-level task $e\in\mathcal{E}$ and a finite horizon $T_{\max}(e)$ measured in environment frames. Let $\mathcal{C}:\mathcal{S}\times\mathcal{E}\rightarrow{0,1}$ denotes the task-completion indicator, where $\mathcal{C}(s_t,e)=1$ if task $e$ has been completed at frame $t$. The episode is successful if
\begin{equation}
 \exists\, t\in\{0,\ldots,T_{\max}(e)\}
 \quad \mathrm{s.t.} \quad
 \mathcal{C}(s_t,e)=1.
\end{equation}
Our objective is to enhance task success ratio across multiple episodes and reduce end-to-end wall-clock runtime.

\section{Method}
\label{sec:method}

\subsection{Overall Framework}
\label{sec:overall}
As illustrated in Figure~\ref{fig:pec_overview}, SyncPlan consists of three components: \textbf{Collaborative Planning} generates low-level action chains for multi-agents; \textbf{Runtime Execution} executes action chains while enforcing synchronization constraints; \textbf{Adaptive Correction} detects plan invalidation to selectively trigger replanning. To enable robust decision-making in complex, dynamic environments, the coordinator is optimized by SFT and RL with a well-designed composite reward function.

\subsection{Collaborative Planning}
\label{sec:plan}
The objective of collaborative planning is to generate long-horizon low-level coordination plans for multi-agents. Different from single-agent planning, multi-agent coordination requires not only deciding what each agent should do, but also considering the collaboration between each agent.
Formally, given a target task $e$ and an environment state $s_t$, the coordinator $\mathcal{D}_\theta$ generates a joint action-chain for $N$ agents.
\begin{equation}
 \Pi_t = \mathcal{D}_\theta(s_t, e) = \{\pi_t^1,\ldots,\pi_t^N\}
 \label{eq:plan-call}
\end{equation}
where $\pi_t^i = [u_{i,1},\ldots,u_{i,L}]$ is the action chain assigned to agent $i$ and $L$ represents the maximum length of action chain. The action space is environment-specific, and the complete action definitions are provided in Appendix A.5.

\textbf{ Explicit Synchronization Primitives}.
Existing work on multi-agent coordination largely relies on implicit cooperation, expressed through multi-turn dialogue \cite{feng2026doctoragent} or simulated communication \cite{zhao2026hicomm}.
In these LLM-based planners, inter-agent coordination dependencies are typically expressed implicitly in natural language (e.g., \textit{"attack after your teammate arrives"}), making them difficult to interpret, verify, and reliably execute during runtime. Consequently, agents normally become unsynchronized, leading to premature actions or coordination failures.
Inspired by process synchronization in operating systems, where coordination is achieved through atomic \texttt{wait} operations, we introduce synchronization primitives that represent coordination as executable runtime constraints.
To be specific, we design two synchronization primitives to capture both inter-agent dependencies and agent–environment interactions:

\begin{itemize}
 \item \textbf{\emph{Wait\_agents}}: agent waits until another agent reaches a specified milestone.
 \item \textbf{\emph{Wait\_entity}}: agent waits until an environment entity satisfies a specified condition.
\end{itemize}
These wait steps make cross-agent dependencies explicit and machine-checkable.

\begin{figure}[!htbp]
\centering
\includegraphics[width=0.45\textwidth]{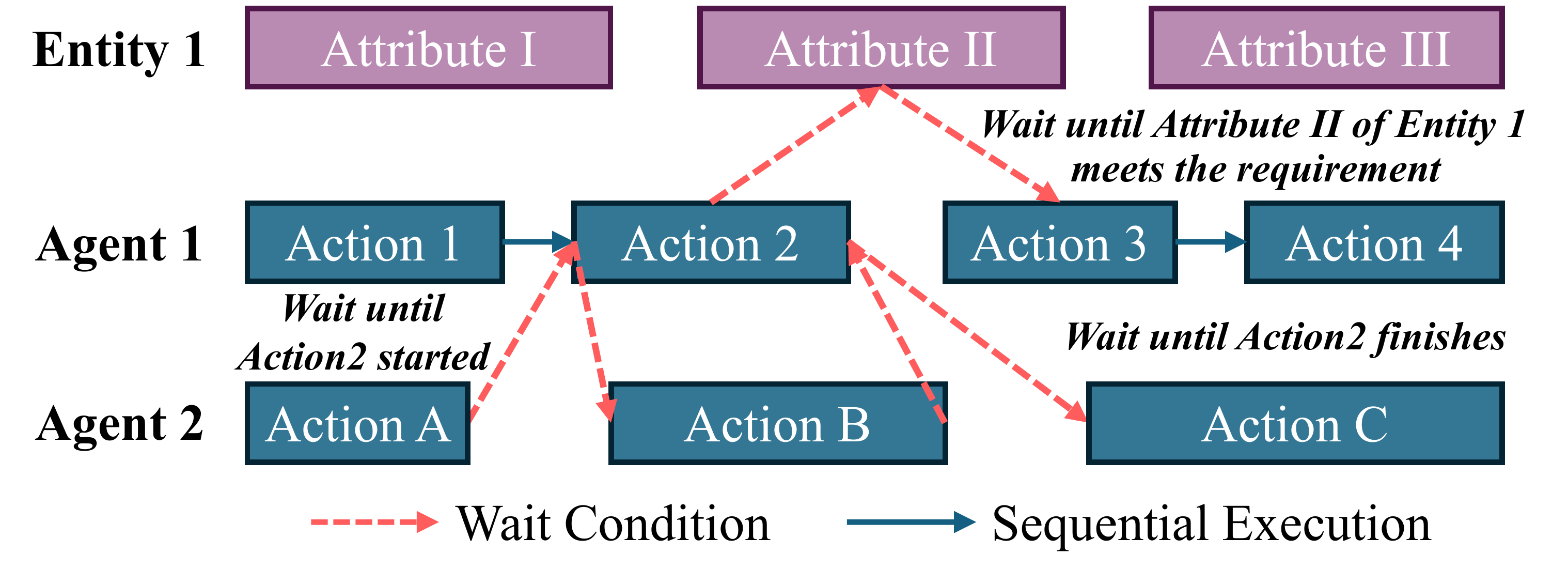}
\caption{\textbf{Runtime execution}. The module advances per-agent action chains and enforces synchronization through agent- and entity-dependent wait conditions.
}
\label{fig:execute_module}
\end{figure}

\subsection{Runtime Execution}
\label{sec:execute}
Considering the generated high-level plan cannot be directly executed by the low-level agents, especially the synchronization primitives, we introduce the execution module to maintain and synchronize the action chain.
This module enables explicit synchronization, as illustrated in Fig.~\ref{fig:execute_module}. A wait step blocks only the corresponding agent, while other agents continue executing their action chains.

Formally, each agent $i$ maintains a pointer $h_i$ to the current head step of $\pi_t^i$. At each frame, the module evaluates the head step $u_{i,h_i}$ according to its type:
\begin{itemize}
 \item \textbf{Executable actions} (e.g., move, attack, interface) are dispatched directly. Once the current action completes, the execution pointer advances to the next action.
 \item \textbf{Synchronization primitives} are interpreted as executable wait predicates. The agent emits a no-op until the corresponding predicate is satisfied.
\end{itemize}
Once the head action is completed, $h_i$ advance to the next action.
This mechanism allows the coordinator to express executable coordination dependencies, such as waiting for a teammate to arrive, waiting for an enemy to become visible, or waiting for an objective to reach a desired state.

\paragraph{Deadlock Detection.}

Considering wait primitives will arise deadlocks, we therefore perform deadlock detection at every frame and trigger the coordinator to replan whenever a deadlock is detected.
Wait-agent dependencies define a directed wait graph $\mathcal{G}_t=(\mathcal{V},\mathcal{E}_t)$, where each node is an agent and an edge $i \rightarrow j$ means that agent $i$ is currently waiting for agent $j$.
Since a circular wait dependency implies that no agent in the cycle can proceed, deadlocks are detected by searching for directed cycles in the wait graph at every frame.
The detailed detection algorithm is provided in Appendix A.4.

\subsection{Adaptive Correction}
\label{sec:acd}
The LLM coordinator assumes that the generated action chains remain valid.
However, in dynamic multi-agent environments, this assumption is frequently violated as evolving game states continuously.
When environmental changes, blindly following the original plan degrades coordination, whereas replanning at every timestep incurs prohibitive computational overhead. Therefore, SyncPlan introduces a lightweight Plan Staleness Detector (PSD) that explicitly determines whether the current collaborative plan remains valid under the latest environment state, triggering replanning only when necessary.

\begin{figure}[!htbp]
 \centering
 \includegraphics[width=0.475\textwidth]{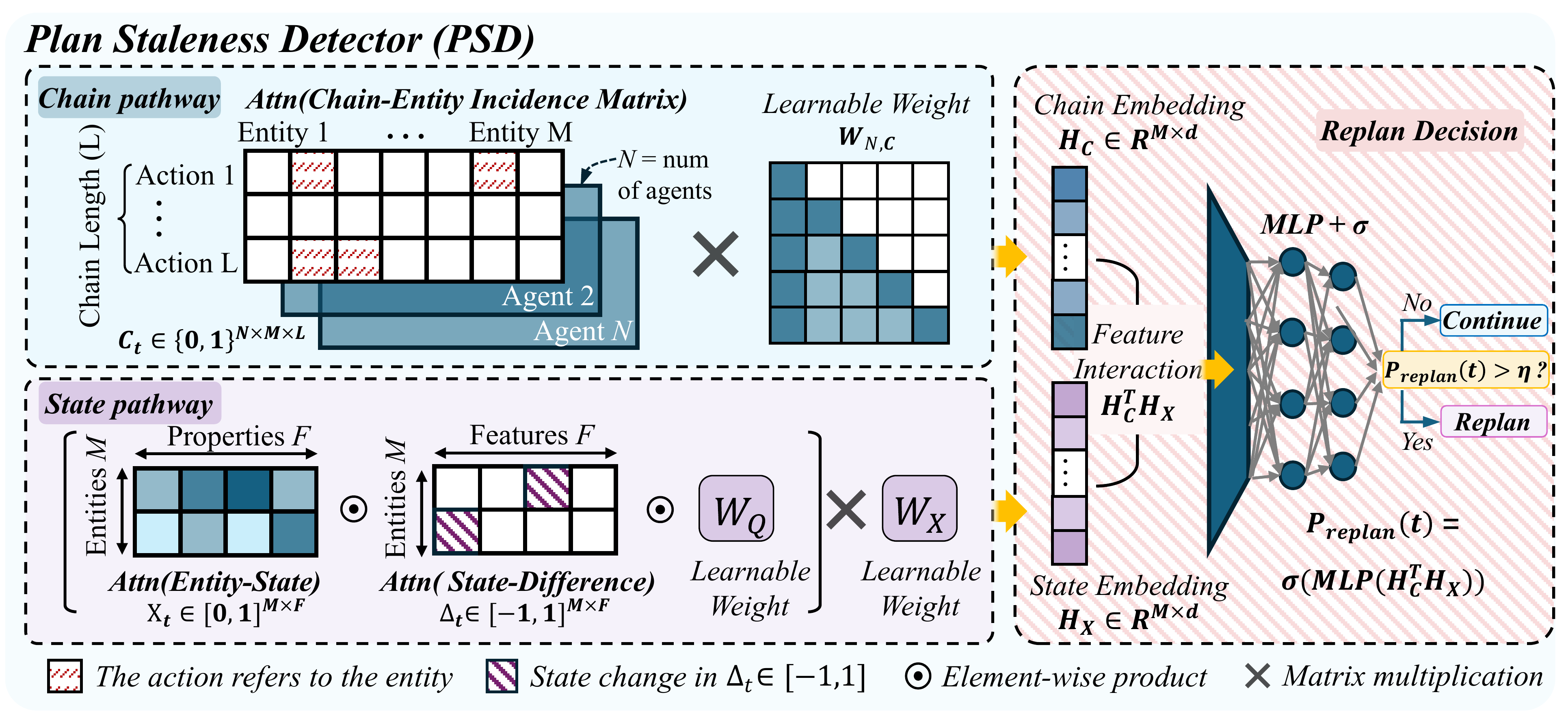}
 \caption{\textbf{Architecture of PSD}. The chain and state pathways encode remaining-plan structure and entity-state changes, whose interaction produces the replanning probability.
 }
 \label{fig:acd_arch}
\end{figure}

At frame $t$, the PSD takes three inputs: (1) the binary chain-entity incidence matrix, which is used to label the participant entity in the action chains, (2) the normalized entity states and (3) the states difference between environments evolving.
To be specific, the inputs are:
\begin{enumerate}
 \item The chain-entity incidence matrix $\mathbf{C}_t \in \{0,1\}^{N\times M \times L}$, where $M$ means the number of entity and $L$ means the max length of one action chain.
 \item The current entity-state matrix $\mathbf{X}_t \in [0,1]^{M \times F}$, where $F$ means the property number of an entity.
 \item The state-difference matrix $\boldsymbol{\Delta}_t \in [-1,1]^{M \times F}$.
\end{enumerate}
The PSD uses two bilinear pathways. The chain pathway embeds which entities are referenced by action chains:
\begin{equation}
 \mathbf{H}_C = \mathbf{W}_{N}^{\{1\times N\}}\times Attn(\mathbf{C}_t) \times\mathbf{W}_C^{\{L\times d\}} \in \mathbb{R}^{M \times d}
\end{equation}
\textbf{Attn()} means the attention module. The state pathway embeds state-conditioned changes:
\begin{equation}
 \mathbf{H}_X = \bigl(Attn(\mathbf{X}_t) \odot Attn(\boldsymbol{\Delta}_t) \odot \mathbf{W}_Q\bigr) \mathbf{W}_X \in \mathbb{R}^{M \times d}
\end{equation}
where $\mathbf{W}_{N,C,Q,W}$ are learnable weight. The two pathways are combined and transformed through MLP module:
\begin{equation}
 P_{\mathrm{replan}}(t) = \sigma \left( MLP\left(\mathbf{H}_C^{\top} \mathbf{H}_X\right)\right)
\end{equation}
A replan is triggered when $P_{\mathrm{replan}}$ exceeds a threshold. For a clear explanation, the architecture is illustrated in Figure~\ref{fig:acd_arch}.

\begin{table*}[!htbp]
\centering
\setlength{\tabcolsep}{3pt}
\caption{Main comparison on Overcooked across standard and dynamic settings. \colorbox[rgb]{ .91, .792, .776}{\textbf{Best}} and \colorbox[rgb]{ .808, .831, .878}{ \underline {second-best}} are highlighted. }
\label{tab:overcooked_main}
\resizebox{\linewidth}{!}{
 \begin{tabular}{cccccccccccccccccc}
 \toprule
 \multirow{2}[4]{*}{Type} & \multicolumn{1}{c}{\multirow{2}[4]{*}{Methods}} & \multicolumn{4}{c}{Standard (Coordination)} & \multicolumn{4}{c}{Dynamic(Coordination)} & \multicolumn{4}{c}{Dynamic(Ring)} & \multicolumn{4}{c}{Dynamic(symmetric)} \\
\cmidrule(lr){3-6}\cmidrule(lr){7-10}\cmidrule(lr){11-14}\cmidrule(lr){15-18} & \multicolumn{1}{c}{} & \multicolumn{1}{c}{TAR $\uparrow$} & AED $\downarrow$ & AAS $\downarrow$ & \multicolumn{1}{c}{RT $\downarrow$ } & \multicolumn{1}{c}{TAR $\uparrow$} & AED $\downarrow$ & AAS $\downarrow$ & \multicolumn{1}{c}{RT $\downarrow$ } & TAR $\uparrow$ & AED $\downarrow$ & AAS $\downarrow$ & \multicolumn{1}{c}{RT $\downarrow$ } & \multicolumn{1}{c}{TAR $\uparrow$} & AED $\downarrow$ & AAS $\downarrow$ & \multicolumn{1}{c}{RT $\downarrow$ } \\
 \midrule
 \multirow{3}[2]{*}{\rotatebox{90}{RL}} & \multicolumn{1}{c}{Greedy} & \multicolumn{1}{c}{40.6\%} & 70.52 & 73.40 & \multicolumn{1}{c}{0.34s} & \multicolumn{1}{c}{58.5\%} & 76.32 & 86.57 & \multicolumn{1}{c}{0.40s} & 22.6\% & 74.08 & 94.91 & \multicolumn{1}{c}{0.30s} & \multicolumn{1}{c}{96.2\%} & 51.20 & 53.08 & \multicolumn{1}{c}{0.26s} \\
 & \multicolumn{1}{c}{PBT} & \multicolumn{1}{c}{92.0\%} & 50.91 & 50.91 & \multicolumn{1}{c}{0.23s} & \multicolumn{1}{c}{88.7\%} & 53.09 & 58.51 & \multicolumn{1}{c}{0.27s} & \cellcolor[rgb]{ .91, .792, .776}\textbf{96.2\%} & \cellcolor[rgb]{ .808, .831, .878}52.78 & \cellcolor[rgb]{ .808, .831, .878}64.60 & \multicolumn{1}{c}{0.26s} & \multicolumn{1}{c}{\cellcolor[rgb]{ .808, .831, .878}98.1\%} & 44.92 & 45.98 & \multicolumn{1}{c}{0.22s} \\
 & \multicolumn{1}{c}{FCP} & \multicolumn{1}{c}{91.0\%} & 72.77 & 75.43 & \multicolumn{1}{c}{1.15s} & \multicolumn{1}{c}{74.0\%} & 75.59 & 82.30 & \multicolumn{1}{c}{1.27s} & \cellcolor[rgb]{ .808, .831, .878}91.0\% & 62.94 & 66.53 & \multicolumn{1}{c}{1.27s} & \multicolumn{1}{c}{94.3\%} & 47.58 & 50.60 & \multicolumn{1}{c}{0.24s} \\
 \midrule
 \multirow{8}[2]{*}{\rotatebox{90}{LLM-Based}} & \multicolumn{1}{c}{A-ToM(1st)} & \multicolumn{1}{c}{74.0\%} & 43.74 & 46.96 & \multicolumn{1}{c}{3.15h} & \multicolumn{1}{c}{60.6\%} & 72.05 & 83.45 & \multicolumn{1}{c}{4.85h} & 67.0\% & 67.50 & 78.67 & \multicolumn{1}{c}{5.36h} & \multicolumn{1}{c}{94.1\%} & 73.25 & 74.88 & \multicolumn{1}{c}{5.82h} \\
 & \multicolumn{1}{c}{A-ToM(2nd)} & \multicolumn{1}{c}{86.0\%} & 44.67 & 44.46 & \multicolumn{1}{c}{3.52h} & \multicolumn{1}{c}{62.3\%} & 65.39 & 78.83 & \multicolumn{1}{c}{6.09h} & 76.0\% & 63.74 & 74.35 & \multicolumn{1}{c}{6.46h} & \multicolumn{1}{c}{94.3\%} & 62.12 & 64.32 & \multicolumn{1}{c}{7.05h} \\
 & \multicolumn{1}{c}{Collab} & \multicolumn{1}{c}{79.5\%} & 55.28 & 64.12 & \multicolumn{1}{c}{5.52h} & \multicolumn{1}{c}{63.0\%} & \cellcolor[rgb]{ .808, .831, .878}53.52 & 70.45 & \multicolumn{1}{c}{6.37h} & 50.7\% & 72.46 & 86.53 & \multicolumn{1}{c}{5.25h} & \multicolumn{1}{c}{95.9\%} & 45.00 & 47.30 & \multicolumn{1}{c}{6.05h} \\
 & \multicolumn{1}{c}{ProAgent} & \multicolumn{1}{c}{88.0\%} & 75.93 & 78.88 & \multicolumn{1}{c}{12.72m} & \multicolumn{1}{c}{65.0\%} & \cellcolor[rgb]{ .808, .831, .878}77.91 & 86.06 & \multicolumn{1}{c}{23.33m} & 59.0\% & 65.50 & 80.12 & \multicolumn{1}{c}{23.40m} & \multicolumn{1}{c}{94.0\%} & 45.56 & 48.82 & \multicolumn{1}{c}{13.91m} \\
 & \multicolumn{1}{c}{DPT-Agent} & \multicolumn{1}{c}{91.3\%} & 67.45 & 74.14 & \multicolumn{1}{c}{38.15m} & \multicolumn{1}{c}{71.5\%} & 63.14 & 74.01 & \multicolumn{1}{c}{53.33m} & 69.3\% & 54.73 & 68.46 & \multicolumn{1}{c}{42.43m} & \multicolumn{1}{c}{96.3\%} & 57.20 & 64.21 & \multicolumn{1}{c}{1.15h} \\
 & \multicolumn{1}{c}{Ours (w/o SFT)} & \multicolumn{1}{c}{92.2\%} & 49.88 & \cellcolor[rgb]{ .808, .831, .878}43.82 & \multicolumn{1}{c}{11.21s} & \multicolumn{1}{c}{88.4\%} & 53.86 & 48.12 & \multicolumn{1}{c}{11.04s} & 76.5\% & 58.80 & 68.91 & \multicolumn{1}{c}{25.26s} & \multicolumn{1}{c}{\cellcolor[rgb]{ .91, .792, .776}\textbf{100\%}} & 49.24 & 36.71 & \multicolumn{1}{c}{17.54s} \\
 & \multicolumn{1}{c}{Ours (w/o RL)} & \multicolumn{1}{c}{\cellcolor[rgb]{ .808, .831, .878}94.8\%} & \cellcolor[rgb]{ .808, .831, .878}43.27 & 43.85 & \multicolumn{1}{c}{6.46s} & \multicolumn{1}{c}{\cellcolor[rgb]{ .808, .831, .878}88.8\%} & 54.13 & 52.58 & \multicolumn{1}{c}{9.55s} & 76.9\% & 68.80 & 76.23 & \multicolumn{1}{c}{6.86s} & \multicolumn{1}{c}{92.2\%} & \cellcolor[rgb]{ .808, .831, .878}40.68 & \cellcolor[rgb]{ .808, .831, .878}32.79 & \multicolumn{1}{c}{5.38s} \\
& \multicolumn{1}{c}{Ours (Full)} & \multicolumn{1}{c}{\cellcolor[rgb]{ .91, .792, .776}\textbf{96.1\%**}} & \cellcolor[rgb]{ .91, .792, .776}\textbf{43.10} & \cellcolor[rgb]{ .91, .792, .776}\textbf{41.88} & \multicolumn{1}{c}{4.91s} & \multicolumn{1}{c}{\cellcolor[rgb]{ .91, .792, .776}\textbf{93.0\%**}} & \cellcolor[rgb]{ .91, .792, .776}\textbf{53.57} & \cellcolor[rgb]{ .91, .792, .776}\textbf{52.09} & \multicolumn{1}{c}{8.24s} & 88.3\%** & \cellcolor[rgb]{ .91, .792, .776}\textbf{52.23} & \cellcolor[rgb]{ .91, .792, .776}\textbf{63.71} & \multicolumn{1}{c}{8.80s} & \multicolumn{1}{c}{\cellcolor[rgb]{ .91, .792, .776}\textbf{100\%**}} & \cellcolor[rgb]{ .91, .792, .776}\textbf{40.18} & \cellcolor[rgb]{ .91, .792, .776}\textbf{27.94} & \multicolumn{1}{c}{4.32s} \\
 \bottomrule
 \end{tabular}%
 }
  \scalebox{0.85}{* p < 0.05, ** p < 0.01, compared with the baseline LLM-based methods under the same setting using a two-sided paired permutation test.
}

\end{table*}

\subsection{SFT and Reinforcement Learning}
\label{sec:sft-rl}
While we can employ a general-purpose LLM as the coordinator, prompting-based SyncPlan still faces two bottlenecks: \textit{deployment latency}, and a \textit{decision quality ceiling} bounded by pre-trained knowledge.
Therefore, we adopt an SFT and RL training pipeline to achieve both real-time responsiveness and superior performance on smaller models.

\subsubsection{SFT Warm-Start}
SFT initializes the coordinator to generate structured, executable multi-agent action chains. We construct instruction-response pairs from expert gameplay trajectories and distilled coordination data, with each target containing complete action chains and synchronization primitives. This establishes a strong prior over the SyncPlan schema and provides in-distribution initialization for subsequent RL. Training details are provided in Appendix B.5.

\subsubsection{Planning-oriented Reinforcement Learning}
After data fine-tuning, the LLM coordinator has developed the ability to coordinate multiple agents. However, it does not explicitly optimize long-horizon execution outcomes, as learning remains driven by token-level imitation rather than trajectory-level execution feedback. Therefore, we formulate executable collaborative planning as a reinforcement learning problem, improving the model's planning quality and reasoning capability by rewarding correct and goal-directed executable action chains, thereby progressively aligning planning intent with executor behavior. We design a composite reward that jointly optimizes three complementary properties of executable collaborative planning:
\begin{itemize}
 \item \textbf{Task Effectiveness}, encouraging successful completion of long-horizon collaborative objectives;
 \item \textbf{Execution Correctness}, discouraging syntax errors, semantic inconsistencies, and deadlocks;
 \item \textbf{Runtime Efficiency}, encouraging efficient coordination with minimal unnecessary waiting and replanning.
\end{itemize}

Let $A$ denote an action chain produced by the coordinator. Executing $A$ in the environment terminates in exactly one of five mutually exclusive outcomes: \textsc{Success}, \textsc{SyntaxError}, \textsc{SemanticError}, \textsc{Deadlock}, and \textsc{Timeout}.
We denote the terminal outcome by $\Omega(A)\in\mathcal{O}$, where
$\mathcal{O}=\{\textsc{Succ},\textsc{Syn},\textsc{Sem},\textsc{Dlk},\textsc{Tmo}\}$.

The reward assigned to $A$ combines a positive \emph{progress} term with a set of outcome-dependent penalties:

\begin{equation}
\resizebox{0.3\textwidth}{!}{$\displaystyle
\begin{aligned}
 R(A) &= \underbrace{R_{\mathrm{prog}}(A) + \alpha\,\mathbb{1}[\Omega(A){=}\textsc{Succ}]}_{\text{intrinsic reward } R_{\mathrm{intr}}(A)}
 \\
 &\quad - \underbrace{(P_{\mathrm{tmo}} + P_{\mathrm{syn}} + P_{\mathrm{sem}} + P_{\mathrm{dlk}})}_{\text{penalty } \mathcal{P}(A)}
\end{aligned}
$}
\label{eq:reward}
\end{equation}

where each penalty term is activated only when the corresponding outcome or event occurs during the execution of $A$:
Intuitively, $R_{\mathrm{prog}}(A)$ measures how much of the task the chain actually accomplishes before termination, and the bonus $\alpha$ is granted only when the chain fully completes the task. The four penalty coefficients $\lambda_{\mathrm{tmo}},\lambda_{\mathrm{syn}},\lambda_{\mathrm{sem}},\lambda_{\mathrm{dlk}}$ discourage the four failure modes with different severity: syntax and deadlock errors are penalized more heavily since they typically indicate that the chain is unusable, while semantic errors and timeouts receive milder penalties as they may still reflect partial progress. The detailed computation  and the specific values of $\alpha,\lambda_{(\cdot)}$ are provided in Appendix A.8.

\begin{table}[!t]
 \centering
 \setlength{\fboxsep}{2pt}
    \caption{Performance comparison on Honor of Kings.}
 \label{tab:main_hok}
 \resizebox{\linewidth}{!}{
 \begin{tabular}{@{}lcccc@{}}
 \toprule
 \multirow{2}{*}{Method}
 & \multicolumn{2}{c}{Gemini-3.5-Flash}
 & \multicolumn{2}{c}{DeepSeek-R1 (Thinking)} \\
 \cmidrule(lr){2-3}\cmidrule(lr){4-5}
 & TAR $\uparrow$ & RTR $\downarrow$
 & TAR $\uparrow$ & RTR $\downarrow$ \\
 \midrule

 \multicolumn{5}{@{}l}{\textit{Off-the-shelf coordinators (w/o SFT)}} \\[1pt]

 \quad ProAgent
 & 56.3 ±3.5\%
 & 15.34 ±1.6x
 & 62.5 ±2.6\%
 & 19.29 ±2.0x \\

 \quad DPT-Agent
 & 68.7 ±2.9\%
 & 21.76 ±2.1x
 & 71.4 ±4.8\%
 & 23.10 ±2.8x \\

 \quad SyncPlan (w/o SFT)
 & \textbf{72.1 ±7.6\%}
 & \textbf{7.32 ±2.0x}
 & \textbf{74.4 ±4.1\%}
 & \textbf{9.41 ±2.5x} \\

 \midrule

 \multicolumn{1}{@{}l}{\textit{Fine-tuned SyncPlan}}
 & \multicolumn{2}{c}{TAR $\uparrow$}
 & \multicolumn{2}{c}{RTR $\downarrow$} \\
 \cmidrule(lr){2-3}\cmidrule(lr){4-5}

 \quad SyncPlan (SFT)
 & \multicolumn{2}{c}{
 \smash{\colorbox[rgb]{.808,.831,.878}{
 \makebox[1.55cm][c]{79.0 ±2.5\%}
 }}
 }
 & \multicolumn{2}{c}{
 \smash{\colorbox[rgb]{.808,.831,.878}{
 \makebox[1.55cm][c]{0.96 ±0.03x}
 }}
 } \\

 \quad SyncPlan (SFT + RL)
 & \multicolumn{2}{c}{
 \smash{\colorbox[rgb]{.91,.792,.776}{
 \makebox[1.55cm][c]{\textbf{86.3 ±5.3\%}}
 }}
 }
 & \multicolumn{2}{c}{
 \smash{\colorbox[rgb]{.91,.792,.776}{
 \makebox[1.55cm][c]{\textbf{0.83 ±0.07x}}
 }}
 } \\
 \bottomrule
 \end{tabular}
 }
\end{table}

\section{Experiments}
\label{sec:experiments}

\subsection{Experimental Setup}
\label{sec:setup}
We evaluate SyncPlan on two complementary multi-agent platforms:

\textbf{Overcooked-AI} is a public two-player cooperative benchmark in which two agents prepare and deliver an onion soup. We evaluate three map layouts with distinct spatial structures: \textbf{\emph{Coordination}}, \textbf{\emph{Ring}}, and \textbf{\emph{Symmetric}}. Under the standard setting, episodes follow the original game dynamics without externally injected perturbations, enabling direct comparison with prior methods. To further evaluate robustness to unexpected environmental changes, we introduce three scripted perturbations: a held-item drop, partial scattering of pot contents, and agent displacement. Detailed layout configurations and event protocols are provided in Appendix B.4.

\textbf{Honor of Kings (HoK)} is a 5v5 MOBA environment with ten agents across two opposing teams. In Commander mode, SyncPlan coordinates up to five allied heroes to eliminate a designated enemy hero, while uncontrolled heroes continue acting autonomously under built-in policies. HoK therefore provides a naturally dynamic testbed for long-horizon multi-agent coordination.

\paragraph{Baselines and HoK disclosure.}
For Overcooked, we compare SyncPlan with task-specific RL methods, including FCP~\cite{strouse2021collaborating}, PBT~\cite{jaderberg2017population}, and Greedy~\cite{carroll2019utility,li2023cooperative}, as well as LLM-based coordinators, including Collab~\cite{sun-etal-2025-collab}, Adaptive-ToM~\cite{mu2026adaptive}, ProAgent~\cite{proagent2024}, and DPT-Agent~\cite{zhang2025leveraging}. For HoK, we focus on LLM-based coordinators sharing the same high-level planning interface, as a task-specific MARL would require a separate training pipeline with extensive environment interaction, bespoke reward design, and difficult joint exploration.

\paragraph{LLM base model and Datasets.}
We evaluate our method on ten representative LLMs: Qwen\cite{hui2024qwen2}, LLaMA \cite{grattafiori2024llama}, GLM \cite{glm5team2026glm5vibecodingagentic}, Claude, GPT \cite{achiam2023gpt}, HY3, Gemini \cite{team2023gemini}, Kimi \cite{team2025kimi}, deepseek-r1 \cite{deepseek_r1} and MiniMax \cite{chen2026minimax}.
For Overcooked, our SFT data is from the Gemini \cite{comanici2025gemini} model distillation process. For HoK, our dataset was sampled from anonymized records of real game matches in the Commander Mode (HoK), where neither user identifiers nor any personally identifiable information was collected to safeguard player privacy.

\paragraph{Metrics.}
We evaluate our approach using Task Achievement Rate (TAR) $\uparrow$ across both environments. In Overcooked, we additionally report Average Episode Duration (AED) $\downarrow$, Average Action Steps (AAS) $\downarrow$, Real-Time (RT) $\downarrow$, and Deadlock Rate $\downarrow$ to measure execution efficiency and coordination failures. In HoK, we measure Real-Time Ratio (RTR) $\downarrow$ and Action Timeout Rate (ATR) $\downarrow$ to quantify computational efficiency and instruction execution reliability. Detailed definitions are provided in appendix B.3.

\begin{table*}[!htbp]
 \centering
  \caption{Ablation study on Overcooked across training regimes and architecture variants on the Coordination layout.}
\resizebox{\linewidth}{!}{
    \begin{tabular}{cccccccccc}
    \toprule
    \multirow{2}[4]{*}{Module} & \multirow{2}[4]{*}{Variant} & \multicolumn{4}{c}{Dynamic}   & \multicolumn{4}{c}{Standard} \\
\cmidrule(lr){3-6}\cmidrule(lr){7-10}         &       & TAR $\uparrow$ & AED $\downarrow$  & AAS $\downarrow$  & RT $\downarrow$   & TAR $\uparrow$ & AED $\downarrow$  & AAS $\downarrow$  & RT $\downarrow$   \\
    \midrule
    \multirow{4}[2]{*}{\rotatebox{90}{w/o SFT}} & w/o PSD &  82.3\% ± 9.5\%   &  50.93 ± 0.6 &  49.29 ± 2.0 &  15.08 ± 1.6 &  92.0\% ± 2.1\%  &  49.88 ± 0.4  &  43.82 ± 0.3  &  11.21 ± 0.3  \\
          & RePlan 4FPS &  64.7\% ± 11.9\%  &  59.27 ± 1.2 &  58.76 ± 2.5 &  91.26 ± 2.9 &  88.2\% ± 3.7\%   &  58.53 ± 1.0 &  56.18 ± 1.8 &  86.48 ± 1.9  \\
          & RePlan 8FPS &  76.5\% ± 10.6\%  &  58.77 ± 1.9 &  59.18 ± 2.9 &  63.60 ± 3.2 &  83.2\% ± 3.1\%   &  54.53 ± 0.5 &  51.47 ± 1.5 &  54.91 ± 2.1  \\
          & SyncPlan &  88.2\% ± 8.1\%   &  53.67 ± 2.3 &  49.29 ± 2.6 &  18.32 ± 2.8 &  92.0\% ± 2.1\%  &  49.88 ± 0.4 &  43.82 ± 0.3 &  11.21 ± 0.3  \\
    \midrule
    \multirow{4}[2]{*}{\rotatebox{90}{w/o RL}} & w/o PSD &  81.4\% ± 4.0\%  & \cellcolor[rgb]{ .808,  .831,  .878} 50.71 ± 0.5 & \cellcolor[rgb]{ .91,  .792,  .776}\textbf{ 46.38 ± 0.9} &  10.09 ± 0.3 &  94.0\% ± 1.3\%  &  49.20 ± 0.1 &  43.87 ± 0.1  &  6.46 ± 0.1   \\
          & RePlan 4FPS &  17.5\% ± 3.9\%  &  61.94 ± 1.8 &  59.39 ± 1.0 &  49.34 ± 0.6 &  16.5\% ± 3.8\%   &  64.88 ± 1.8 &  55.19 ± 0.7   &  45.19 ± 0.3  \\
          & RePlan 8FPS &  29.9\% ± 4.7\%  &  71.31 ± 3.0 &  66.91 ± 1.5 &  29.03 ± 0.5 &  14.4\% ± 3.6\%   &  62.57 ± 4.9 &  56.39 ± 0.9   &  26.02 ± 0.3  \\
          & SyncPlan & \cellcolor[rgb]{ .808,  .831,  .878} 88.7\% ± 3.2\%  &  54.53 ± 1.2 &  50.66 ± 1.3  & \cellcolor[rgb]{ .808,  .831,  .878} 8.85 ± 0.3 & \cellcolor[rgb]{ .808,  .831,  .878} 94.0\% ± 1.3\%  & \cellcolor[rgb]{ .91,  .792,  .776}\textbf{ 49.20 ± 0.1} & \cellcolor[rgb]{ .808,  .831,  .878} 43.87 ± 0.1 & \cellcolor[rgb]{ .808, .831, .878} 6.46 ± 0.1   \\
    \midrule
    \multirow{4}[2]{*}{\rotatebox{90}{SyncPlan}} & w/o PSD &  82.3\% ± 9.5\%   &  50.79 ± 0.6 &  48.24 ± 2.2  &  11.15 ± 0.7 &  96.0\% ± 2.0\%  &  49.53 ± 0.2 &  41.88 ± 0.3 &  4.91 ± 0.1  \\
          & RePlan 4FPS &  23.5\% ± 10.6\%  &  62.00 ± 3.1 &  62.06 ± 1.9 &  41.50 ± 3.3 &  17.6\% ± 9.5\%   &  63.33 ± 5.4 &  53.71 ± 1.2  &  23.68 ± 0.4  \\
          & RePlan 8FPS &  23.5\% ± 10.6\%  &  56.50 ± 3.2 &  64.29 ± 3.7 &  17.41 ± 0.8 &  29.4\% ±5.4\%  &  71.80 ± 9.0 &  57.88 ± 2.0   &  16.34 ± 0.6  \\
          & SyncPlan & \cellcolor[rgb]{ .91,  .792,  .776}\textbf{ 94.1\% ± 5.9\%  } & \cellcolor[rgb]{ .91,  .792,  .776}\textbf{ 50.25 ± 3.0} & \cellcolor[rgb]{ .808,  .831,  .878} 48.24 ± 2.6 & \cellcolor[rgb]{ .91,  .792,  .776}\textbf{ 6.58 ± 0.5} & \cellcolor[rgb]{ .91,  .792,  .776}\textbf{ 96.0\% ± 2.0\% } & \cellcolor[rgb]{ .808,  .831,  .878} 49.53 ± 0.2 & \cellcolor[rgb]{ .91,  .792,  .776}\textbf{ 41.88 ± 0.3} & \cellcolor[rgb]{ .91,  .792,  .776}\textbf{ 4.91 ± 0.0 } \\
    \bottomrule
    \end{tabular}%
 }
   \label{tab:ablation}
\end{table*}

\begin{figure*}[!htbp]
 \centering
 \includegraphics[width=\linewidth]{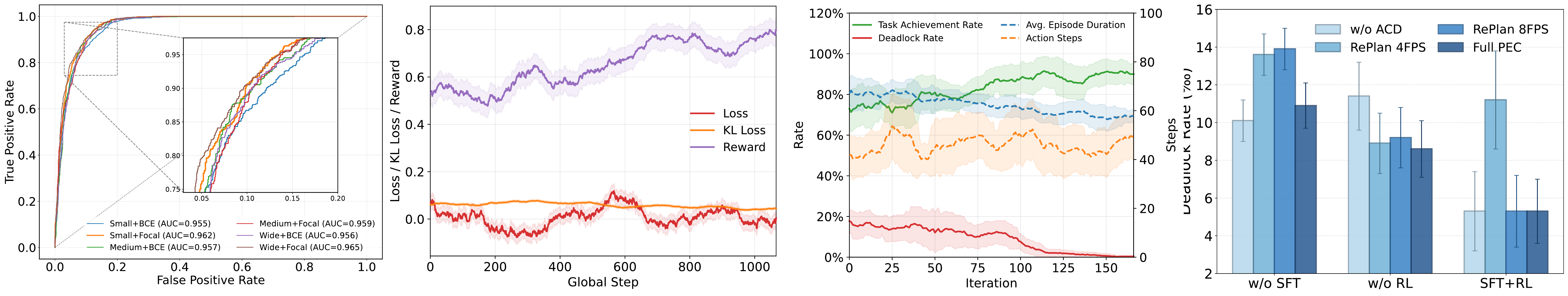}
 \caption{ (A) The ROC curve of PSD model. (B) RL training curves. (C) Task performance during RL. (D) Deadlock ratios.}
 \label{fig:train}
\end{figure*}

\subsection{Main Comparison}
\label{sec:main_comparison}
We benchmark SyncPlan against five external baselines with three architecture-stripped SyncPlan variants under a unified Overcooked protocol covering one Standard and three dynamic evaluation tasks (Table~\ref{tab:overcooked_main}).

The central finding is that SyncPlan substantially outperforms existing LLM-based coordinators on dynamic Overcooked tasks while maintaining much lower wall-clock latency. Compared with task-specific MARL baselines, SyncPlan achieves higher TAR on three settings.
This advantage stems from SyncPlan's event-driven correction design: instead of re-querying the LLM at a fixed frequency or per frame, it invokes the coordinator only when the PSD detects plan invalidation, reducing wall-clock overhead by two to three orders of magnitude compared with A-ToM and Collab.
On the challenging \emph{Dynamic Ring} layout, SyncPlan reaches 88.3\% TAR, exceeding the LLM-based baselines A-ToM and Collab, while remaining below the best RL baseline.

In HoK, we compare SyncPlan against two popular baseline methods, with results presented in Table \ref{tab:main_hok}. Our approach demonstrates lower latency by leveraging long-horizon planning. More importantly, SyncPlan and its variants exhibit significantly stronger coordination performance, achieving 86.3\% TAR compared to the best baseline's 68.75\% in
this complex multi-agent environment.

\begin{figure*}[!htbp]
 \centering
 \includegraphics[width=0.92\linewidth]{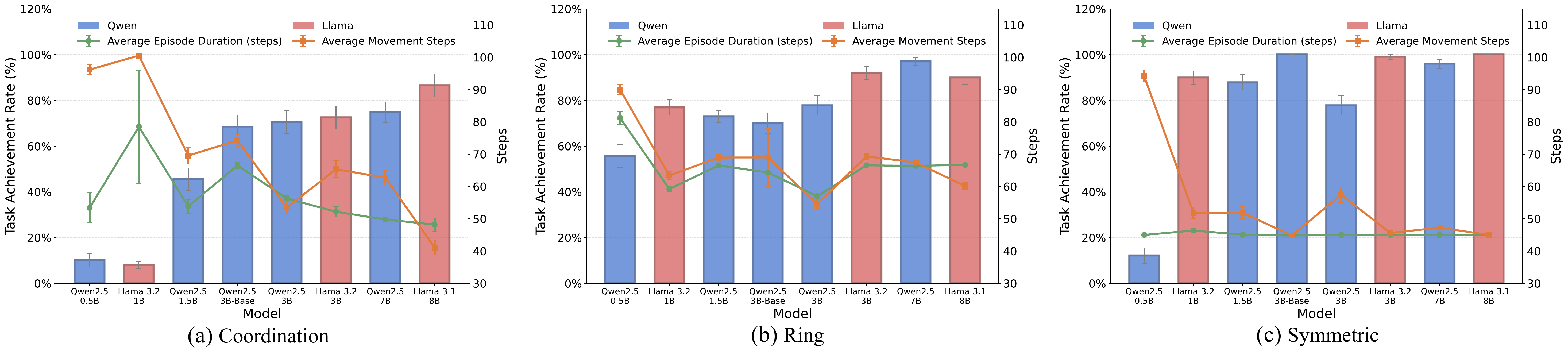}
 \caption{Effect of coordinator model size on SyncPlan performance. The critical threshold lies between 1B and 1.5B.}
 \label{fig:size}
\end{figure*}

\subsection{Ablation Study}
\label{sec:ablation}

Table~\ref{tab:ablation} isolates each architectural choice by comparing Full SyncPlan against w/o PSD and two periodic replanning baselines across three training regimes. Key insights:

\begin{table}[htbp]
 \centering
 \caption{Ablation study on Honor of Kings.}
 \label{tab:hok_ablation}%
\resizebox{\linewidth}{!}{
    \begin{tabular}{ccccc}
    \toprule
    Module & Variant & TAR  $\uparrow$ & ATR $\downarrow$ & RTR  $\downarrow$ \\
    \midrule
    \multirow{4}[2]{*}{w/o SFT} & w/o PSD & 59.0±11.1\% & 8.2±2.1\% & 5.92±1.3x \\
          & RePlan 4FPS & 67.9±4.7\% & \cellcolor[rgb]{ .808,  .831,  .878}0.1±0.0\% & 81.17±7.7x \\
          & RePlan 8FPS & 69.7±6.4\% & 0.1±0.1\% & 57.52±9.4x \\
          & SyncPlan & 72.1±7.6\% & 5.8±1.3\% & 7.32±0.1x \\
    \midrule
    \multirow{4}[2]{*}{w/o RL} & w/o PSD & 66.7±3.3\% & 7.9±4.0\% & \cellcolor[rgb]{ .808,  .831,  .878}0.72±0.0x \\
          & RePlan 4FPS & 60.0±6.9\% & \cellcolor[rgb]{ .91,  .792,  .776}\textbf{0.0±0.0\%} & 2.38±0.1x \\
          & RePlan 8FPS & 66.7±4.8\% & 0.1±0.1\% & 1.04±0.0x \\
          & SyncPlan & 79.0±2.5\% & 3.9±3.4\% & 0.96±0.0x \\
    \midrule
    \multirow{4}[2]{*}{SFT+RL} & w/o PSD & 71.4±8.4\% & 6.6±1.5\% & 0.72±0.0x \\
          & RePlan 4FPS & 73.0±7.1\% & \cellcolor[rgb]{ .91,  .792,  .776}\textbf{0.0±0.0\%} & 2.18±0.1x \\
          & RePlan 8FPS & \cellcolor[rgb]{ .808,  .831,  .878}77.1±9.4\% & 0.1±0.1\% & 1.19±0.0x \\
          & SyncPlan & \cellcolor[rgb]{ .91,  .792,  .776}\textbf{86.3±5.3\%} & 2.3±1.6\% & \cellcolor[rgb]{ .91,  .792,  .776}\textbf{0.83±0.0x} \\
    \bottomrule
    \end{tabular}%
 }
\end{table}%

\textbf{PSD provides targeted, low-cost correction.} On the standard layout, removing PSD has no effect because no dynamic perturbation ever invalidates the initial plan. Under dynamic events, the gap widens sharply: PSD recovers broken plans exactly when needed, whereas the w/o PSD variant cannot react and loses 8+ TAR points. Periodic RePlan baselines attempt to compensate by blind re-invocation, but they pay an order-of-magnitude RT penalty for marginal or even negative TAR gains, because frequent unnecessary replanning destabilizes partially executed chains. This confirms that \emph{when} to correct matters more than \emph{how often}.
\begin{figure}[!htbp]
 \centering
 \includegraphics[width=\linewidth]{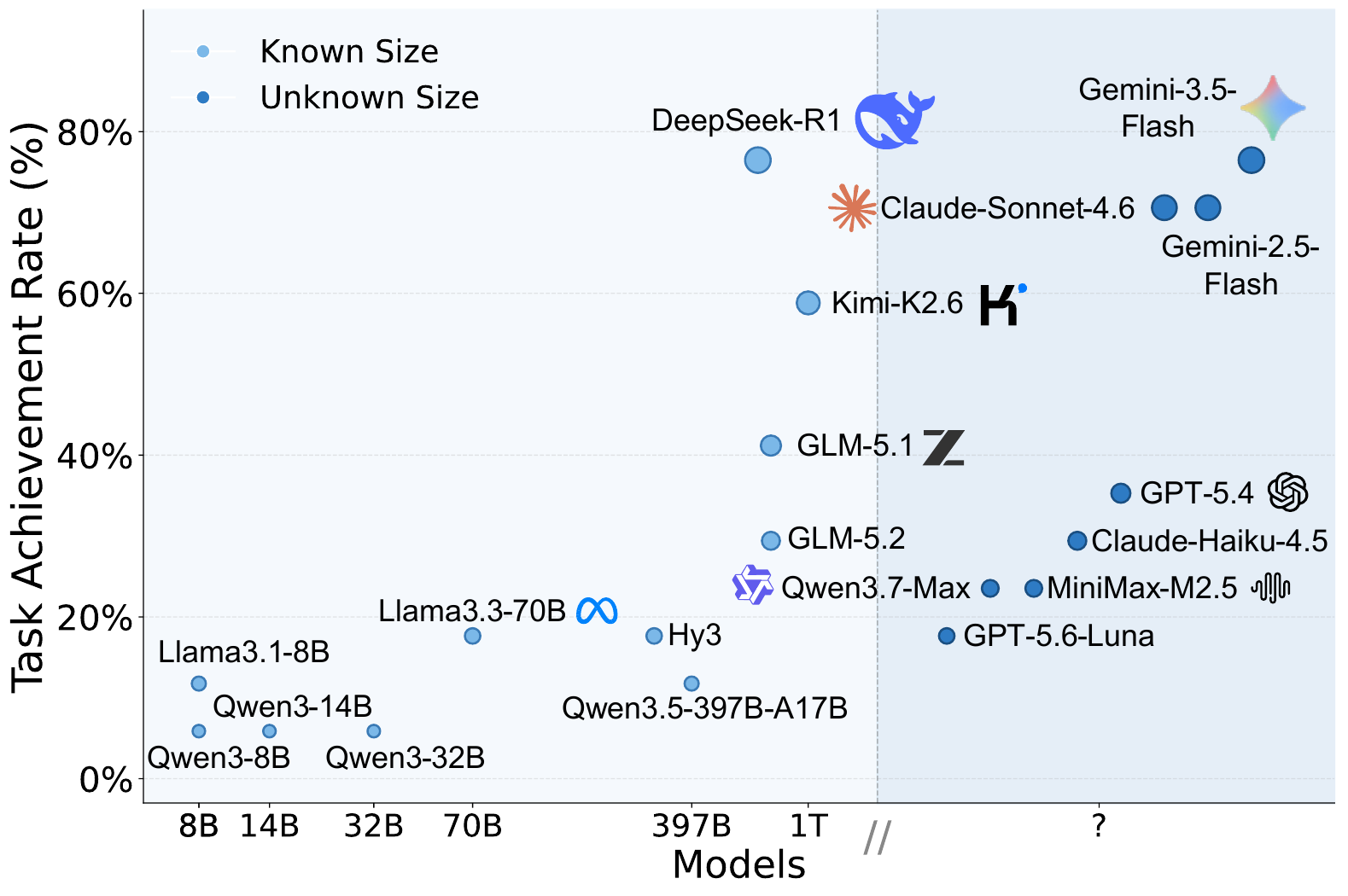}
 \caption{Performance on main backbone models.}
 \label{fig:model_cap}
\end{figure}
\textbf{RL improves coordination quality and boosts TAR.} Without RL (middle block), even Full SyncPlan achieves only 88.7\% dynamic TAR because the SFT-only coordinator, despite generating syntactically correct plans, fails to account for the bot executor's actual movement timing and collision dynamics, leading to poor coordination quality. RL enhances coordination by providing closed-loop feedback that helps the model learn execution-aligned strategies absent from offline demonstrations.
The HoK ablation (Table~\ref{tab:hok_ablation}) confirms these insights: SFT+RL Full SyncPlan reaches 86.3\% TAR versus 79.0\% without RL, while reducing ATR from 7.9\% to 2.3\%. These improvements demonstrate that RL refines the coordinator's ability to generate plans that align with real execution constraints, thereby improving overall coordination quality and task success.

\subsection{Training and Performance Analysis on base models}
\label{sec:behavior}
Figure~\ref{fig:train} reveals how the training stages shape coordinator behavior in complementary ways.
Fig.~\ref{fig:train}(A) shows the ROC curves, demonstrating that PSD performs well under different MLP architectures.
Fig.~\ref{fig:train} (B,C) shows progressively improved TAR by exposing the coordinator to real execution feedback: the reward curve climbs steadily as the model learns to account for bot-level timing constraints that are invisible in the static SFT corpus. A byproduct is the deadlock ratio trajectory.
Fig.~\ref{fig:train}D shows Full SyncPlan maintains the lowest deadlock density throughout training, while w/o SFT exhibits higher rates. This shows deadlock suppression stems from both training stages: SFT teaches structurally valid action chains that avoid resource conflicts, while RL further reduces deadlocks through execution-level collision feedback.
We report the performance of backbone models in Fig.~\ref{fig:model_cap}, from which we observe that Gemini-3.5-Flash and DeepSeek-R1 achieve the leading performance.

\begin{figure}[!htbp]
 \centering
 \includegraphics[width=\linewidth]{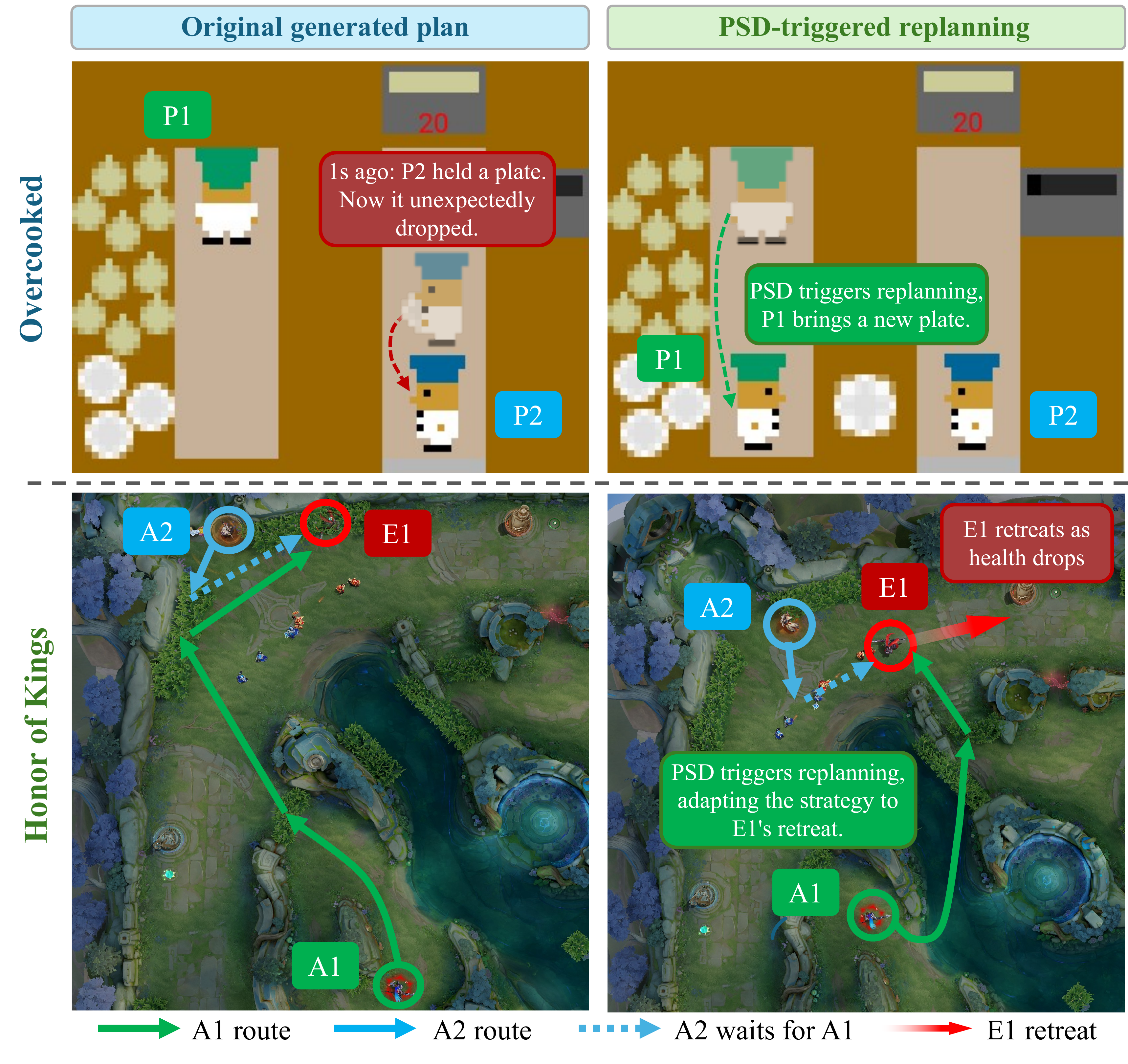}
 \caption{PSD-triggered replanning in two scenarios.}
 \label{fig:case}
\end{figure}
\subsection{Boundary Exploration}
\label{sec:scaling}
To investigate the relationship between model capacity and planning performance, and to identify the minimum parameter threshold required for reliable action generation under structured constraints.
We evaluate LLaMA-3.2-1B/3B, LLaMA-3.1-8B, and Qwen2.5-(0.5/1.5/3/7)B-Instruct. Performance (Fig.~\ref{fig:size}) scales log-linearly with parameter count, but the critical threshold lies between 1B and 1.5B: models below 1B fail to produce syntactically valid action chains at non-trivial rates, making downstream execution unreliable regardless of the framework. Above 3B, gains plateau under SyncPlan's structured schema because the schema itself constrains the output space, meaning additional capacity yields diminishing returns once format compliance is achieved.

\subsection{Case Studies}
\label{sec:casestudy}
Figure~\ref{fig:case} illustrates how SyncPlan's modules collaborate in two scenarios. When PSD detects that current plan is invalidated (P2 drops a plate, E1's health is fast dropping). The PSD detects it and invokes LLM coordinator to update agent's action chain to adjust the changes of  environment's state. This reactive correction completes in limited wall-clock time, whereas a fixed-interval replanner would either react too late or waste compute on unnecessary reinvocations.
\section{Conclusion}
\label{sec:conclusion}
Aiming at coordinating multi-agents in real-time and dynamic environments, we presented \textbf{\emph{SyncPlan}} to separate planning from execution and real-time correction detection. \textbf{\emph{SyncPlan}} contains a synchronization-based LLM coordinator and an action correction detector to maintain the validity of plan.
Experiments across different domains supported that the explicit synchronization enables effective coordination. This study also reveals that RL fine-tuning is critical for bridging LLM coordinator intent and reality across different domains.
Future work will explore SyncPlan’s scalability, generalization to new environments, and larger team sizes. Additionally, we will enhance its recovery mechanisms against execution failures, persistent stalls, and rapid environmental dynamics.

\bibliography{references}

\clearpage
\appendix

\section{Method Details}

\subsection{Algorithm}
\label{sec:algorithm}

Algorithm~\ref{alg:pec} separates the slow planning path from the per-frame execution path. The coordinator $\mathcal{D}_\theta$  first generates an initial multi-agent plan and initializes chain pointers and the wait graph for execution tracking (line 2). At each timestep, the system executes the current plan step and checks for deadlock situations (lines 4-5). If the task is completed, the algorithm terminates successfully (lines 6-8). Otherwise, the Plan Staleness Detector (PSD) evaluates whether replanning is necessary by analyzing the current context, execution status, and state changes (lines 9-10). Replanning is triggered either when a deadlock is detected or when the staleness probability exceeds the threshold $\eta$ (line 11), in which case a new plan is generated and execution trackers are reset (lines 12-13). If the plan remains valid, only the chain pointers are updated to reflect execution progress (line 15). This design enables efficient plan reuse while maintaining adaptive coordination through selective replanning.

\subsection{Training Pipeline Overview}
\label{app:training_pipeline}
\begin{figure*}[htbp]
    \centering
    \includegraphics[width=1\linewidth]{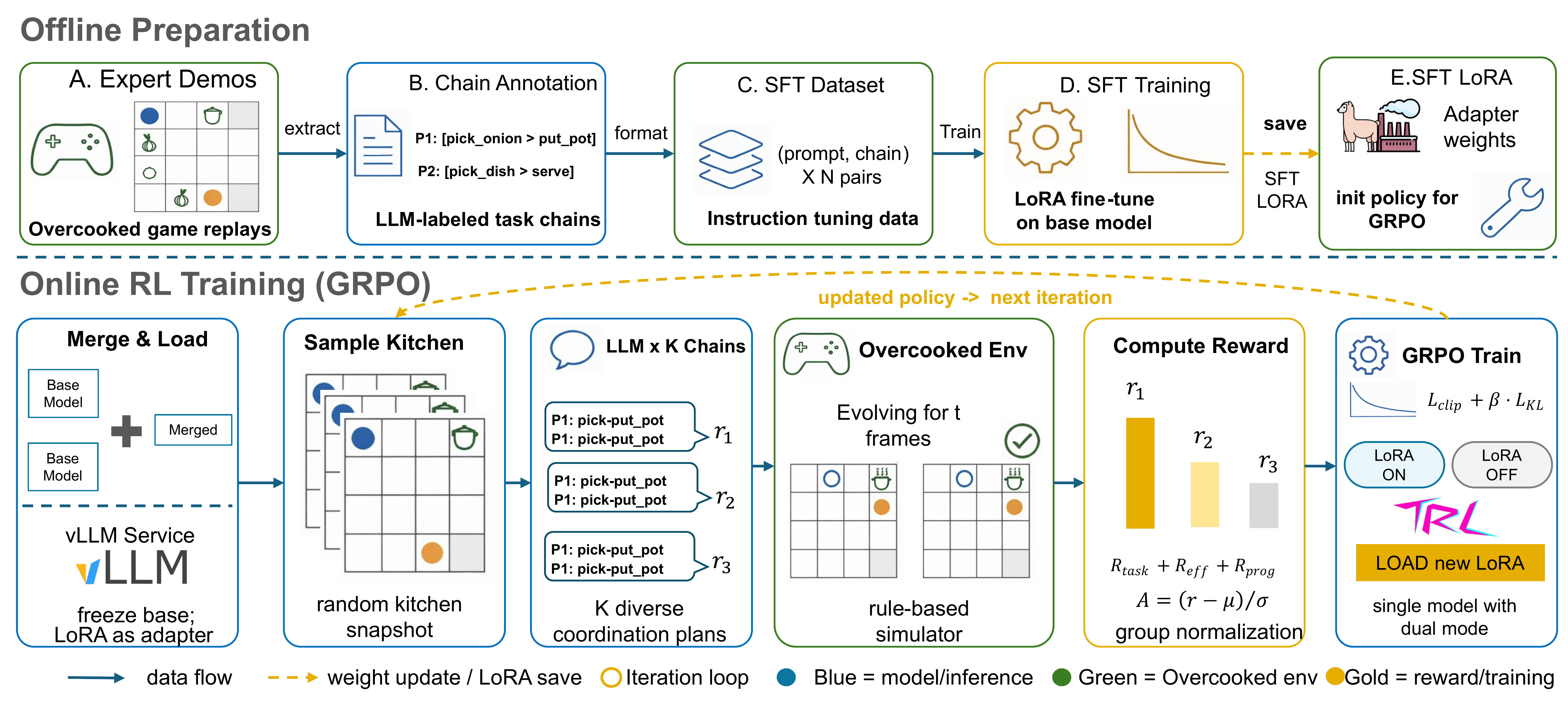}
    \caption{Overview of the two-stage SFT$+$RL training pipeline for the SyncPlan Coordinator. Stage~1 performs supervised fine-tuning; Stage~2 applies GRPO reinforcement learning with environment-grounded rewards.}
    \label{fig:sft_rl}
\end{figure*}

Figure~\ref{fig:sft_rl} illustrates the two-stage training pipeline used to optimize the SyncPlan Coordinator. Stage~1 (SFT) initializes the Coordinator with behavioral priors. Stage~2 (GRPO) further refines the policy through reinforcement learning with the composite reward signal defined in Table~\ref{tab:reward_config}.
We train the Coordinator using LoRA adapters in both stages. We first collect gameplay data and annotate it with labels. We use this labeled data to perform SFT, which produces the first-generation LoRA adapter. We then merge this adapter into the base model to obtain the SFT model. This model also serves as the starting point for the RL stage.

At the beginning of the RL stage, we initialize the training loop with a LoRA adapter whose weights are all zero. This adapter does not change the behavior of the base model. It only provides a general interface for loading LoRA weights during the loop. During the RL loop, we use random seeds to generate different game matches, and we disable scripted dynamic events in the environment. As a result, RL does not directly train on injected perturbations; instead, the gains observed in dynamic evaluation arise from better alignment between the Coordinator and the real executor under the environment's timing and collision constraints. We separate inference and training by using vLLM for inference and TRL for training.

\begin{algorithm}[t]
\caption{SyncPlan}
\label{alg:pec}
\begin{algorithmic}[1]
\REQUIRE task $e$, initial state $s_0$, correction threshold $\eta$
\STATE $\Pi \leftarrow \mathcal{D}_\theta(s_0,e)$
\STATE initialize chain pointers $\{h_i\}$ and wait graph $\mathcal{G}$
\FOR{each frame $t$}
    \STATE $\mathbf{a}_t,\mathrm{deadlock} \leftarrow
    \textsc{ExecuteStep}(\Pi_t,\{h_i\},\mathcal{G},s_t)$
    \STATE apply $\mathbf{a}_t$ and observe $s_{t+1}$
    \IF{$\mathcal{C}(s_{t+1},e)=1$}
        \STATE \textbf{return} \textsc{Success}
    \ENDIF
    \STATE construct $\mathbf{C}_{t+1},\mathbf{X}_{t+1},\boldsymbol{\Delta}_{t+1}$
    \STATE $P_{\mathrm{replan}} \leftarrow
    \mathrm{PSD}(\mathbf{C}_{t+1},\mathbf{X}_{t+1},\boldsymbol{\Delta}_{t+1})$
    \IF{$\mathrm{deadlock}$ or $P_{\mathrm{replan}}>\eta$}
        \STATE $\Pi \leftarrow \mathcal{D}_\theta(s_{t+1},e)$
        \STATE reset $\{h_i\}$ and $\mathcal{G}$
    \ELSE
        \STATE Update $\{h_i\}$
    \ENDIF
\ENDFOR
\end{algorithmic}
\end{algorithm}

\subsection{Complete Prompt Templates}
\label{app:prompts}
This subsection provides the full prompt templates used in the SyncPlan Plan Module. The Plan Module is invoked as a single structured LLM call that produces per-agent action chain assignments. We use \textbf{two prompt variants} depending on whether the underlying model has been fine-tuned on the target environment: a \textit{Non-SFT} variant for off-the-shelf base/instruct models, and a \textit{SFTed} variant for models that have already internalized the action vocabulary through supervised fine-tuning or RL on SyncPlan trajectories.

\paragraph{Prompt schema overview.}
Both variants follow the same three-role chat layout. The \textbf{System} message encodes the static task framing: the environment \textit{Background}, the coordination \textit{Target}, and the number of controlled agents. The \textbf{User} message supplies the dynamic \textit{Current Environment State} at the current decision step. The \textbf{Assistant} response emits per-agent action chains inside tagged blocks, one block per agent. The variants differ only in whether the action vocabulary (\textit{Description} and \textit{Usage}) is materialized in the System message.

\paragraph{Variant 1: Non-SFT prompt.}
For base or instruct models with no environment-specific fine-tuning, the System message must spell out the full action vocabulary so the model can ground its outputs:

\begin{small}
\begin{verbatim}
[System]  You are a planner for {Background}.
          Your target is {Target} while
          controlling {Num of Agents} agents.
          You assign each agent an action
          chain. The action list is
          {Description}; the usage is {Usage}.
          There are some cases: {Case}.

[User]    Here is the current state: {State}
[Assistant]
          {"agent_1":{action1, action2, ...},
           "agent_2":{action1, action2, ...}
          ...
          }
\end{verbatim}
\end{small}

\paragraph{Variant 2: SFTed prompt.}
For models that have been fine-tuned on SyncPlan trajectories from the target environment, the action vocabulary is already internalized in the weights, so the Description and Usage slots are dropped from the System message.
\begin{small}
\begin{verbatim}
[System]  You are a planner for {Background}.
          Your target is {Target} while
          controlling {Num of Agents} agents.
[User]    Here is the current state: {State}
[Assistant]
          {"agent_1":{action1, action2, ...},
           "agent_2":{action1, action2, ...}
          ...
          }
\end{verbatim}
\end{small}

\noindent Slots in braces (\textit{Background}, \textit{Target}, \textit{Num of Agents}, \textit{Description}, \textit{Usage}, \textit{State}) are template variables filled at planning time; tokens inside the \texttt{Assistant} blocks are generated by the model.

\subsection{Deadlock Detection}
\label{app:deadlock}

This detector addresses \emph{agent-dependency deadlocks}: cycles induced by active \texttt{Wait\_agents} predicates. It does not classify an unsatisfied \texttt{Wait\_entity} predicate, a timeout, or another non-cyclic execution failure as a graph deadlock; those cases remain available to the PSD or the episode-level termination logic. We maintain a directed dependency graph $\mathcal{G} = (\mathcal{V}, \mathcal{E})$, where $\mathcal{V} = \{1, \ldots, K\}$ and an edge $i \!\to\! j \in \mathcal{E}$ exists only while agent $i$ is blocked by an active $\texttt{Wait\_agents}(j, m, s)$ predicate. A graph deadlock corresponds to a directed cycle in $\mathcal{G}$. When agent $i$ enters a new wait $\texttt{Wait\_agents}(j, m, s)$, we test whether $j$ can already reach $i$ in $\mathcal{G}$ \emph{before} inserting the edge:
\begin{equation}
    \mathcal{R}(j, i) \;=\;
    \begin{cases}
        \text{true} & \text{if } i = j, \\
        \text{true} & \text{if } \exists k.\ (j,k) \!\in\! \mathcal{E} \,\land\, \mathcal{R}(k, i), \\
        \text{false} & \text{otherwise}.
    \end{cases}
    \label{eq:reach}
\end{equation}
If $\mathcal{R}(j, i) = \text{true}$, inserting $i \!\to\! j$ would close a cycle, which means a deadlock is found. Otherwise, the edge is appended to $\mathcal{E}$. When agent $i$'s wait is satisfied, the edge $i \!\to\! j$ is removed in $O(1)$; since the remaining graph stays acyclic by construction, no rebuild is required. Each reachability check is bounded by $O(|\mathcal{V}| + |\mathcal{E}|) = O(K^2)$ in the worst case, but in practice each agent has at most one outgoing wait edge ($|\mathcal{E}| \!\le\! K$), reducing the cost to $O(K)$.

\subsection{The design details of action chains}

\label{app:intent_space}
In this section, we introduce the details of action chains.
\subsubsection{Action-Chain Step Language}
\label{sec:step-language}
Each step-level objective $o_i^{(\ell)}$ takes one of two forms: an action or a wait condition. This design follows the basic distinction between execution and synchronization in concurrent programming. An action primitive specifies an operation that can be verified by the environment. A wait primitive specifies a condition on other agents or on the world state that must be satisfied before the next step can proceed. This design gives the LLM an explicit way to represent synchronization, rather than forcing it to hide such logic in natural-language step descriptions.
{\footnotesize
\begin{equation*}
\begin{aligned}
\langle\text{step}\rangle & ::= \langle\text{action}\rangle \mid \langle\text{wait}\rangle \\
\langle\text{wait}\rangle & ::= \langle\text{atom}\rangle \mid \langle\text{atom}\rangle\ (\texttt{\&}\mid\texttt{|})\ \langle\text{atom}\rangle\\\
\langle\text{atom}\rangle & ::= \textbf{\emph{Wait\_agents}}(a, \{\text{start}|\text{done}\}, s) \\
                         & \mid\; \textbf{\emph{Wait\_entity}}(e, \text{attr}, \text{op}, v)
\end{aligned}
\end{equation*}
}
The action vocabulary depends on the environment. In contrast, the wait primitives capture two synchronization patterns that appear in both environments. In \textbf{\emph{Wait\_agents}}(a, m, s), $a$ identifies the prerequisite agent, $m\in\{\textbf{\emph{start}},\textbf{\emph{done}}\}$ selects whether the prerequisite is the start or completion milestone, and $s$ identifies the referenced step in that agent's current chain. The predicate is removed as soon as the indicated milestone is observed. \textbf{\emph{Wait\_entity}} pauses progress until a condition on the world state becomes true, based on an exposed entity attribute such as hero health or pot status. We allow bounded \texttt{\&} and \texttt{|} composition, so the language can express common coordination rules such as "both teammates have arrived" or "low HP or under attack" while keeping each predicate evaluation local to the current frame. This language is used directly by the Execute Module. A \textbf{\emph{Wait\_agents}} cycle triggers the graph-deadlock pathway in Appendix~\ref{app:deadlock}; non-cyclic unsatisfied waits are not silently reclassified as deadlocks.

\subsubsection{The action list of Honor of Kings}
Table~\ref{tab:intent_hok} defines the complete action-primitive vocabulary used by the Plan Module in the HoK environment.

\begin{table}[htbp]
\centering
\footnotesize
\setlength{\tabcolsep}{3pt}
\resizebox{\columnwidth}{!}{%
\begin{tabular}{lll}
\toprule
\textbf{Type} & \textbf{Target} & \textbf{Description} \\
\midrule
\textsc{Kill} & Enemy or monster & Kill the specific target \\
\textsc{Move} & Position & Move to specific position \\
\textsc{Push} & Tower & Destroy the specific tower \\
\textsc{Clear} & None & Kill all the minions around \\
\textsc{Back} & None & Back to home \\
\bottomrule
\end{tabular}%
}
\caption{HoK action-primitive vocabulary. }
\label{tab:intent_hok}
\end{table}
In Honor of Kings, we support the five common action types described above. The LLM is required to coordinate participant agents using these five actions with the two wait primitives.

\subsubsection{The action list of  Overcooked.}
Table~\ref{tab:intent_oc} defines the action vocabulary for Overcooked.

\begin{table}[htbp]
\centering
\footnotesize
\setlength{\tabcolsep}{3pt}
\begin{tabular}{llp{4.5cm}}
\toprule
\textbf{Type} & \textbf{Target} & \textbf{Description} \\
\midrule
\textsc{Interact} & Target & Interact with the specified target \\
\textsc{Move\_to} & Position & Move to specific position \\
\textsc{Face} & Direction & Face the specific direction \\
\bottomrule
\end{tabular}
\caption{Overcooked action-primitive vocabulary. \textit{Target} specifies the station or ingredient; \textit{Position} is the grid coordinate.}
\label{tab:intent_oc}
\end{table}

\subsection{The design details of PSD}
\subsubsection{The design principle}
Since the detector must operate at high frequency to capture situational changes in real time, efficiency is the primary design goal. We therefore use a simple and lightweight structure to track state changes in relevant entities.

\subsubsection{The input of PSD model}
Since the PSD module must monitor both the absolute state and the relative change of every entity in the environment, its input is built from two complementary entity-level tensors rather than raw observations. At each frame $t$, we compute the normalized entity-state matrix $\mathbf{X}_t \in [0,1]^{M \times F}$, which encodes the current absolute state of all $M$ tracked entities across $F$ normalized attributes, and the state-change matrix $\boldsymbol{\Delta}_t \in [-1,1]^{M \times F}$, obtained by differencing $\mathbf{X}_t$ against the state $k_{\mathrm{ref}}$ frames earlier. We use $k_{\mathrm{ref}}=3$ (Table~\ref{tab:reward_config}). Feeding both the absolute state $\mathbf{X}_t$ and the differential signal $\boldsymbol{\Delta}_t$ lets the detector distinguish a stable-but-unfavorable situation from one that is rapidly deteriorating.

In addition to $\mathbf{X}_t$ and $\boldsymbol{\Delta}_t$, we construct the chain-entity incidence tensor $\mathbf{C}_t \in \{0,1\}^{N \times M \times L}$ for the \emph{remaining} plan and $N$ agents. Specifically, $C_{t,i,j,\ell}=1$ if entity $j$ is referenced by the unexecuted step $\ell\ge h_i$ in agent $i$'s current chain; completed-prefix steps are masked out. This tensor supplies a relevance prior: rather than weighting every entity equally, it directs the PSD toward entities referenced by the portion of the plan that can still be invalidated.

\paragraph{Tensor contraction and output shape.}
The concise notation in Equations~(4)--(6) contracts the agent axis of $\mathbf{C}_t$. Concretely, $\mathbf{W}_C\in\mathbb{R}^{L\times d}$ first maps the step axis to $d$ dimensions, and $\mathbf{W}_N$ then sums over controlled agents, yielding $\mathbf{H}_C\in\mathbb{R}^{M\times d}$. For the state pathway, $\mathbf{W}_Q\in\mathbb{R}^{F}$ is broadcast across entities as a feature-wise gate and $\mathbf{W}_X\in\mathbb{R}^{F\times d}$ maps the gated state-difference features to $\mathbf{H}_X\in\mathbb{R}^{M\times d}$. Thus $\mathbf{H}_C^{\top}\mathbf{H}_X\in\mathbb{R}^{d\times d}$; the MLP maps this interaction matrix to one logit, and the sigmoid produces the scalar $P_{\mathrm{replan}}(t)\in(0,1)$. This convention is used consistently in the implementation and in the main-paper equations.
\begin{figure}[htbp]
    \centering
    \includegraphics[width=0.85\linewidth]{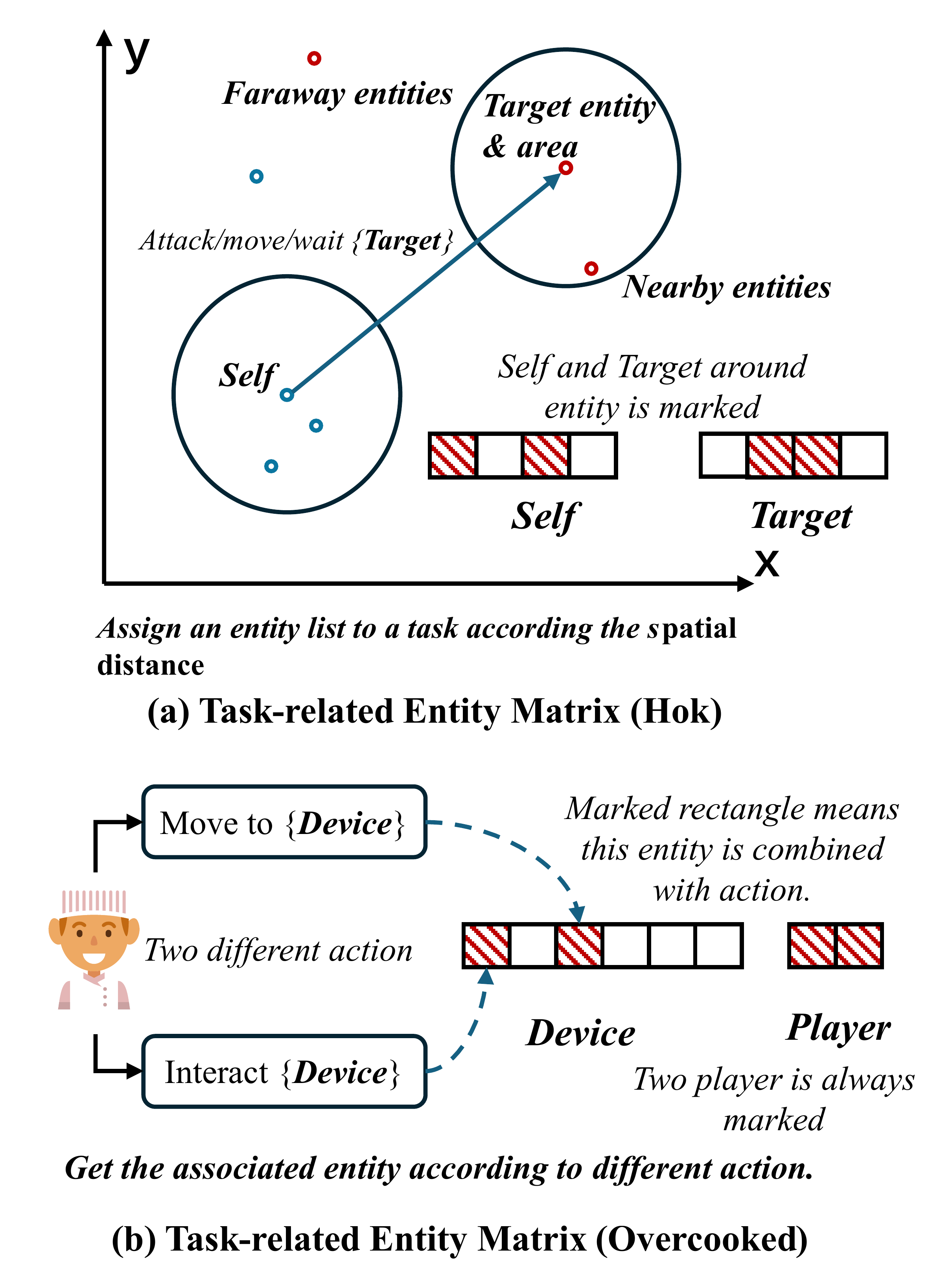}
    \caption{Constructing the remaining-plan chain-entity incidence tensor $\mathbf{C}_t$ per domain. \textbf{(a) Honor of Kings}: spatial-proximity rule. \textbf{(b) Overcooked}: action-object binding.}
    \label{fig:entity_matrix}
\end{figure}

\subsubsection{Action-Chain Incidence Matrix}
\label{app:incidence_matrix}
The definition of entity relevance in $\mathbf{C}_t$ depends on the domain. For this reason, we do not learn $\mathbf{C}_t$ from data. Instead, we construct it with a lightweight heuristic that reflects the structure of each environment. Figure~\ref{fig:entity_matrix} shows the resulting entity relevance matrices in the two domains.

In Honor of Kings, most chain steps are spatial. Typical examples include \textit{Attack}, \textit{Move}, and \textit{Wait} with respect to a target location or target entity. In this setting, whether an entity is relevant mainly depends on spatial proximity. For each step, we always include the self-agent. We also include entities within a fixed radius around the self-agent, as well as entities within a fixed radius around the target location or target entity. All included entities are marked as relevant by setting $C_{t,i,j,\ell}=1$ for the corresponding agent $i$, entity $j$, and unexecuted step $\ell$. Entities that are far away from both the self-agent and the target remain $0$ (Figure~\ref{fig:entity_matrix}a).

In Overcooked, most chain steps are centered on specific objects or agents. Typical examples include \emph{Move\_to} and  \emph{Interact \{Device\}}.
In this setting, relevance is determined by the object referenced in the step template rather than by spatial distance. We parse the operation template of the current step and identify the referenced entity category, such as a \emph{Device} slot or a \emph{Player} slot. The corresponding category is then marked as relevant in $\mathbf{C}_t$, while unrelated devices and players remain $0$ (Figure~\ref{fig:entity_matrix}b). This design keeps $\mathbf{C}_t$ simple and fully rule-based. It does not introduce any additional learnable parameters. It is also easy to extend to a new environment. One only needs to define an appropriate relevance rule based on the structure of that domain.

\subsubsection{Training Data \& Labels}
\label{app:PSD_training}

The PSD is trained as a binary classifier: given the input tensors $(\mathbf{C}_t, \mathbf{X}_t, \boldsymbol{\Delta}_t)$ at frame $t$, it predicts whether the remaining chain should be corrected, supervised by a binary label $y_t \in \{0,1\}$. Here $y_t=1$ denotes an operationally defined plan-staleness event: an observed or constructed change that invalidates an assumption used by at least one remaining chain step; $y_t=0$ denotes a clean frame with no such event. Training data for the two environments is constructed independently from the SFT corpus (Section~\ref{app:sft_data}) and is not shared across environments.

\paragraph{Overcooked: event-frame labeling.} We reuse the dynamic-event injection protocol (Section~\ref{app:oc_events}): the frame at which a dynamic event is injected is labelled $y_t=1$, since the event is designed to invalidate at least one field of the active chain. Frames drawn from clean (non-injected) rollouts are labelled $y_t=0$. Because positive frames are rare relative to the length of a rollout, we subsample negative frames at a positive-to-negative ratio of $1{:}5$ to mitigate the class imbalance that would otherwise bias the classifier toward always predicting $y_t=0$.

\paragraph{Honor of Kings: synthetic perturbation.} HoK has no scripted event injection at \emph{evaluation} time (Section~\ref{app:dyn_events}); however, to obtain a sufficiently dense supervision signal for \emph{training} the PSD, we construct positive examples offline by synthetically perturbing recorded frames so that the active chain becomes invalid. Two perturbation types are used: (i) reducing the health of an allied agent referenced by the active chain, simulating an unexpected engagement risk; and (ii) moving an enemy hero closer to the chain's target location or agent, simulating an unanticipated rotation. Each perturbed frame is manually reviewed to confirm that the perturbation indeed renders the original chain unreasonable before being accepted as a positive example; frames without perturbation are labelled $y_t=0$. This synthetic construction is used exclusively to train the PSD offline and does not alter the injection-free evaluation protocol described in Section~\ref{app:dyn_events}, under which HoK dynamics arise entirely from uncontrolled bot behavior.

\paragraph{Data collection and split.} The HoK and Overcooked PSD training sets are each collected independently of the corresponding SFT rollout pool, and each is split into training and validation subsets with an $8{:}2$ ratio.

\subsubsection{PSD Threshold Selection}
The PSD threshold $\theta_{\text{PSD}}$ trades off between recall (catching genuine chain invalidations) and precision (avoiding unnecessary replanning). We select $\theta_{\text{PSD}}=0.7$ via grid search over $\{0.3, 0.5, 0.7, 0.9\}$ on a held-out validation set, balancing detection recall with replanning overhead.

\begin{figure}[htbp]
\centering
\includegraphics[width=0.9\linewidth]{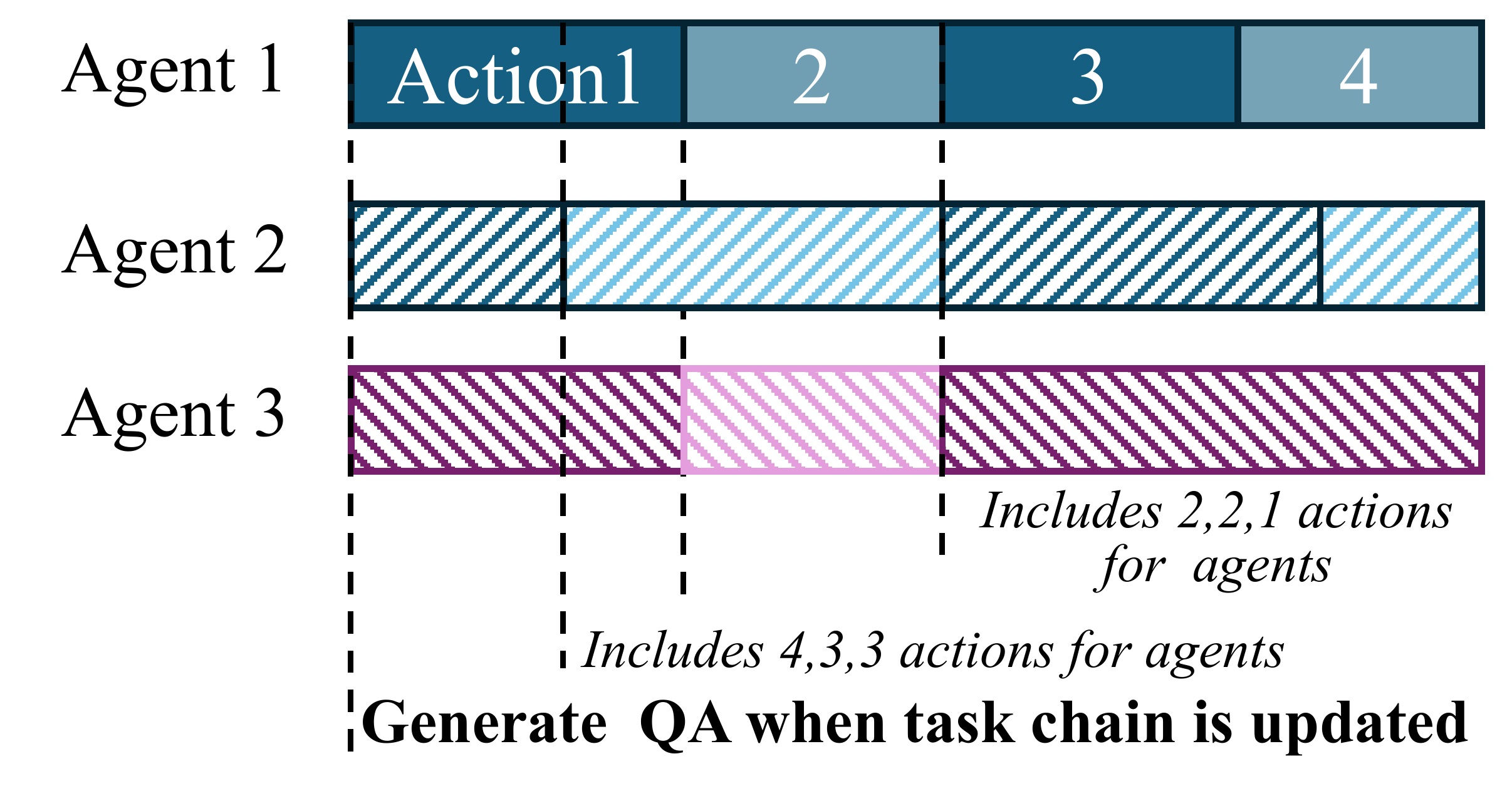}
\caption{Event-driven SFT data augmentation. Each dashed vertical line marks an action-chain update event. Within one segment, different agents may execute different numbers of actions under the same chain. For example, the first segment has counts \texttt{(4,3,3)}, while the next replanned segment has counts \texttt{(2,2,1)}.}
\label{fig:sft_data_aug}
\end{figure}
\subsection{SFT Data Augmentation}
\label{app:sft_data}

Our SyncPlan-RL Coordinator is first warm-started with a short supervised fine-tuning (SFT) stage before online RL, and we also report an SFT-only baseline in the main paper. This section describes how the SFT corpus is constructed. A naive per-step imitation objective is inefficient, because many consecutive frames share the same action chain and therefore provide highly repetitive supervision. In contrast, the informative decision points are the moments when the Coordinator updates the chain. We therefore use an event-driven data augmentation scheme that converts each chain-update event into one QA training instance, while also recording the multi-agent action footprint within the corresponding segment.

\paragraph{Construction protocol.} We replay SyncPlan trajectories collected from the prompt-only Coordinator and treat each chain-update event as a supervision boundary. For each rollout, we apply the following procedure:
\begin{enumerate}
  \item \textbf{Detect chain-update events.} A boundary $t_k$ is recorded whenever the Plan Module emits a new action-chain set $\{\tau_i\}_{t_k}$.

  \item \textbf{Aggregate per-agent action footprints.} For each segment $[t_k, t_{k+1})$, we record, for every controlled agent $i$, how many atomic actions were actually executed under the same chain. For example, the first segment in Figure~\ref{fig:sft_data_aug} has action counts $(4,3,3)$. These counts are stored as auxiliary metadata for each segment.

  \item \textbf{Generate one QA instance per boundary.} For each boundary $t_k$, we create one QA pair. The \emph{question} consists of the Coordinator observation at $t_k$, including the current state $o_{t_k}$, the previous chain set $\{\tau_i\}_{t_{k-1}}$, and the correction history $h_{t_k}$. The \emph{answer} is the new chain set $\{\tau_i\}_{t_k}$ emitted at that boundary, together with the per-agent action-footprint statistics collected from the subsequent segment $[t_k, t_{k+1})$.

\end{enumerate}

\subsection{Reward Design}
\label{app:reward}

This section provides the full definition of the reward $R(A)$ used in GRPO training. Executing a generated action chain terminates in exactly one of five mutually exclusive outcomes: \textsc{Success}, \textsc{SyntaxError}, \textsc{SemanticError}, \textsc{Deadlock}, or \textsc{Timeout}. For an action chain $A$, the reward is
\begin{equation*}
\begin{split}
    R(A) = R_{\mathrm{prog}}(A) + \alpha \, \mathbb{1}[\Omega(A)=\textsc{Success}]\\
    - \bigl(P_{\mathrm{syn}}(A) + P_{\mathrm{sem}}(A) + P_{\mathrm{dlk}}(A) + P_{\mathrm{tmo}}(A)\bigr).
\end{split}
\end{equation*}
Each penalty is zero unless its corresponding terminal outcome occurs. The reward is computed once per response and then clamped to the range $[-1,1]$.

\paragraph{Progress reward.}
The term $R_{\mathrm{prog}}(A)$ combines three dense signals. The first is execution progress within the generated chain, rather than a second task-success indicator:
\begin{equation*}
R_{\mathrm{prog}}(A)
=w_{\mathrm{frac}} R_{\mathrm{frac}}(A)
+w_{\mathrm{eff}} R_{\mathrm{eff}}(A)
+w_{\mathrm{len}} R_{\mathrm{len}}(A).
\end{equation*}
Each component is normalized to $[0,1]$ before weighting.

\begin{itemize}[leftmargin=1.2em, itemsep=2pt]
\item $R_{\mathrm{frac}}(A)$ measures fractional progress toward the action chain:
$$
R_{\mathrm{frac}}(A)=\frac{n_{\text{completed}}}{n_{\text{total}}},
$$
where $n_{\text{completed}}$ is the number of completed actions and $n_{\text{total}}$ is the length of action chain. This term is defined for every response, regardless of the final outcome.

\item $R_{\mathrm{eff}}(A)$ rewards faster successful completion:
$$
R_{\mathrm{eff}}(A)=1-\frac{t_{\text{finish}}}{T}
$$
if the task is completed successfully, and $R_{\mathrm{eff}}(A)=0$ otherwise. Here $t_{\text{finish}}$ is the frame at which the task is completed, and $T$ is the episode horizon.

\item $R_{\mathrm{len}}(A)$ is a length regularization term. It favors responses whose length falls within the ideal token range $[L_{\min},L_{\max}] = [100,500]$, and decreases outside this range, with a maximum reduction of $0.5$.
\end{itemize}

\paragraph{Penalty terms.}
The penalty terms correspond to the four non-success outcomes. With the values in Table~\ref{tab:reward_config}, deadlock receives the largest fixed penalty ($1.0$), semantic failure receives $0.3$, syntax failure receives $0.1$ scaled by the fraction of malformed instructions, and timeout has no additional fixed penalty; a timeout still forfeits both the success bonus and the efficiency reward. This explicit numerical ordering is the one used in training.

\begin{itemize}[leftmargin=1.2em, itemsep=2pt]
\item \textbf{Syntax penalty.}
If the response cannot be parsed, we apply
$$
P_{\mathrm{syn}}(A)=\lambda_{\mathrm{syn}} \cdot \mathrm{sev}_{\mathrm{syn}}(A).
$$
The severity $\mathrm{sev}_{\mathrm{syn}}(A)\in[0,1]$ equals $1$ for a fully unparsable response, such as invalid JSON. For a partially malformed response, it is defined as the fraction of invalid instructions among all instructions.

\item \textbf{Semantic penalty.}
If the response is syntactically valid but fails semantic validation during execution, we apply a fixed penalty
$$
P_{\mathrm{sem}}(A)=\lambda_{\mathrm{sem}}
$$
Typical cases include references to non-existent targets or invalid entities.

\item \textbf{Deadlock penalty.}
If execution enters deadlock, we apply
$$
P_{\mathrm{dlk}}(A)=\lambda_{\mathrm{dlk}}
$$

\item \textbf{Timeout penalty.}
If the response times out without triggering syntax, semantic, or deadlock failure, we set
$$
P_{\mathrm{tmo}}(A)=\lambda_{\mathrm{tmo}}
$$
In this case, the response receives no additional fixed penalty, but it also does not receive the success bonus $\alpha$ or the efficiency reward $R_{\mathrm{eff}}$. This makes timeout the mildest of the four failure modes.
\end{itemize}

\begin{table}[htbp]
\centering
\footnotesize
\begin{tabular}{lc}
\toprule
\textbf{Parameter} & \textbf{Value} \\
\midrule
\multicolumn{2}{l}{\textit{Success bonus}} \\
\quad $\alpha$ & 0.45 \\
\midrule
\multicolumn{2}{l}{\textit{Penalty coefficients}} \\
\quad $\lambda_{\mathrm{syn}}$ (syntax error) & 0.1 \\
\quad $\lambda_{\mathrm{sem}}$ (semantic error) & 0.3 \\
\quad $\lambda_{\mathrm{dlk}}$ (deadlock) & 1.0 \\
\quad $\lambda_{\mathrm{tmo}}$ (timeout) & 0.0 \\
\midrule
\multicolumn{2}{l}{\textit{$R_{\mathrm{prog}}$ internal weights}} \\
\quad $w_{\mathrm{frac}}$ (fractional progress) & 0.15 \\
\quad $w_{\mathrm{eff}}$ (efficiency) & 0.15 \\
\quad $w_{\mathrm{len}}$ (length term) & 0.25 \\
\midrule
\multicolumn{2}{l}{\textit{Environment}} \\
\quad Horizon $T$ & 100 steps \\
\quad $k_{\text{ref}}$ (replan reference) & 3 \\
\midrule
\multicolumn{2}{l}{\textit{Length regularization}} \\
\quad Ideal token range $[L_{\min}, L_{\max}]$ & $[100,500]$ \\
\quad Maximum reduction & 0.5 \\
\bottomrule
\end{tabular}
\caption{Configuration of the GRPO reward function.}
\label{tab:reward_config}
\end{table}

\section{Experiment Details}
\subsection{Computing infrastructure.}
All experiments were conducted on Linux servers running TencentOS Server 4.2. Each server is equipped with dual-socket AMD EPYC 9K84 96-Core processors and 8 NVIDIA H20 GPUs, each with 96\,GB memory. The software environment uses Python 3.11.6, NVIDIA driver 535.161.08, and CUDA 12.8. Our training and inference stack is built on PyTorch 2.10.0, Transformers 4.57.6, TRL 0.24.0, vLLM 0.19.0, Accelerate 1.7.0, DeepSpeed 0.16.4, PEFT 0.18.1, and LLaMA-Factory 0.9.5.dev0.

\subsection{Environment Layouts}
\label{app:overcooked}

\begin{figure}[htbp]
\centering
\includegraphics[width=\columnwidth]{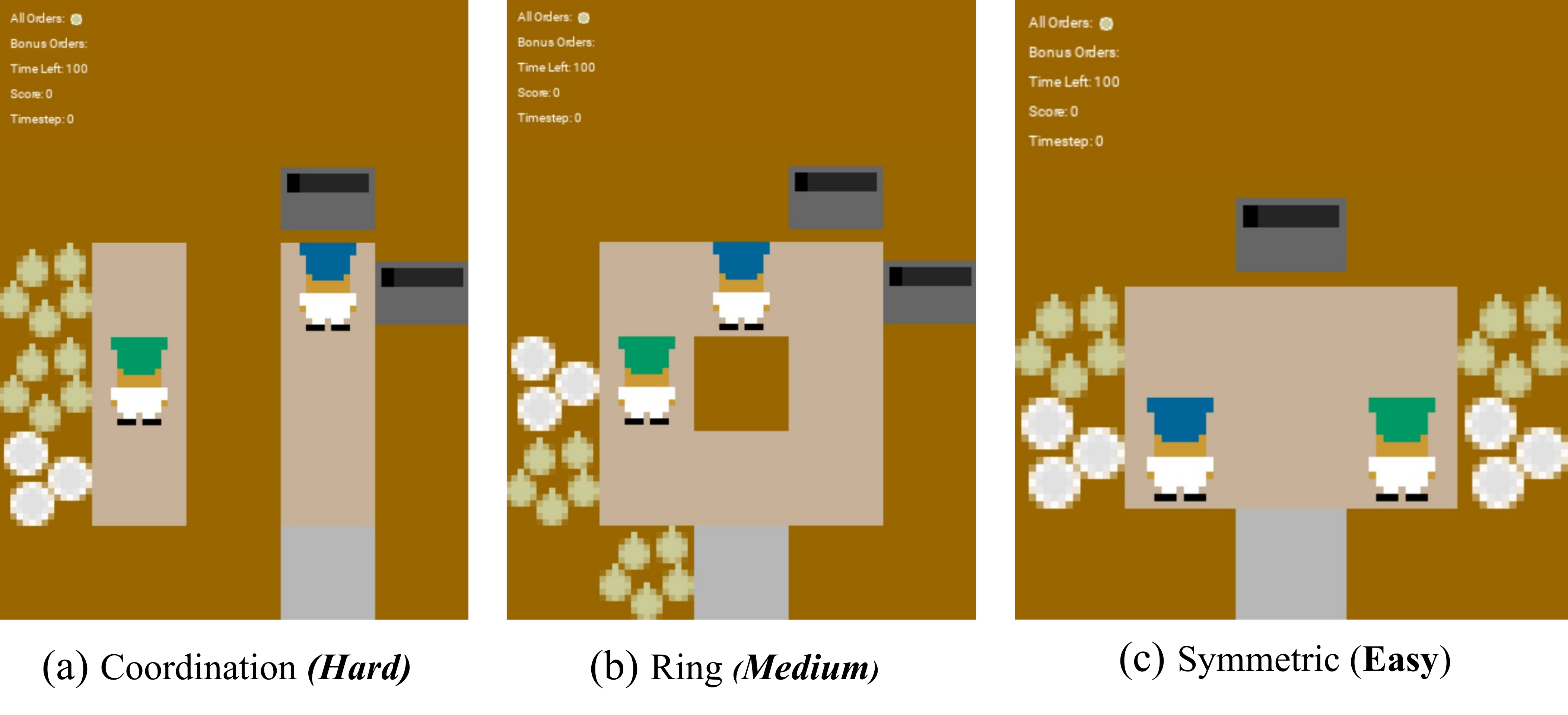}
\caption{Our experiments are conducted on three different layouts.}
\label{fig:layouts}
\end{figure}

We evaluate on three Overcooked layouts shown in Figure~\ref{fig:layouts}: \textit{Coordination}, \textit{Ring}, and \textit{Symmetric}. The standard evaluation uses the original, unperturbed dynamics on the Coordination layout. Dynamic evaluation applies the seeded perturbation protocol in Section~\ref{app:oc_events} to each of the three layouts, yielding the Dynamic (Coordination), Dynamic (Ring), and Dynamic (Symmetric) settings reported in the main comparison.

\subsection{Evaluation metrics}
In this section, we define the reported metrics. For each evaluation setting, we run independent trials and report the mean and standard deviation across trials. The random seed for the $i$-th repeated trial is set to $i$, which ensures reproducible replay under the same evaluation protocol. In HoK, we additionally randomize the controlled heroes' starting positions and orientations before each episode to increase evaluation diversity. HoK starting points are selected from player-match records using an internal filtering pipeline and then replayed under the same seed-controlled setup for all compared methods.

We report four common metrics in both environments.  Task Achievement Ratio (TAR) is the percentage of trials in which the target task is successfully achieved. Average Eposide Duration (AED), Average Action Steps(AAS), and Running Time (RT) denote the average episode duration, average movement steps, and wall-clock running time, respectively, computed over repeated trials.
We also report domain-specific diagnostic metrics. For Overcooked, the deadlock rate is defined as the fraction of executed actions that result in deadlock. For HoK, we report Action Timeout Rate (ATR), defined as the fraction of coordinator action whose execution exceeds the allowed wall-clock budget.

\paragraph{Paired statistical testing.} For the marked main-table comparisons, the $i$-th run of each method is paired through the same evaluation seed and, in Overcooked, the same seeded event realization. We apply a two-sided paired permutation test to the trial-level outcomes under each setting. The significance markers in the main comparison denote comparisons of SyncPlan (SFT+RL) with the LLM-based baselines in the corresponding setting; the markers are not intended as a claim of significance against task-specific RL baselines.

\begin{figure}[htbp]
\centering
\includegraphics[width=\columnwidth]{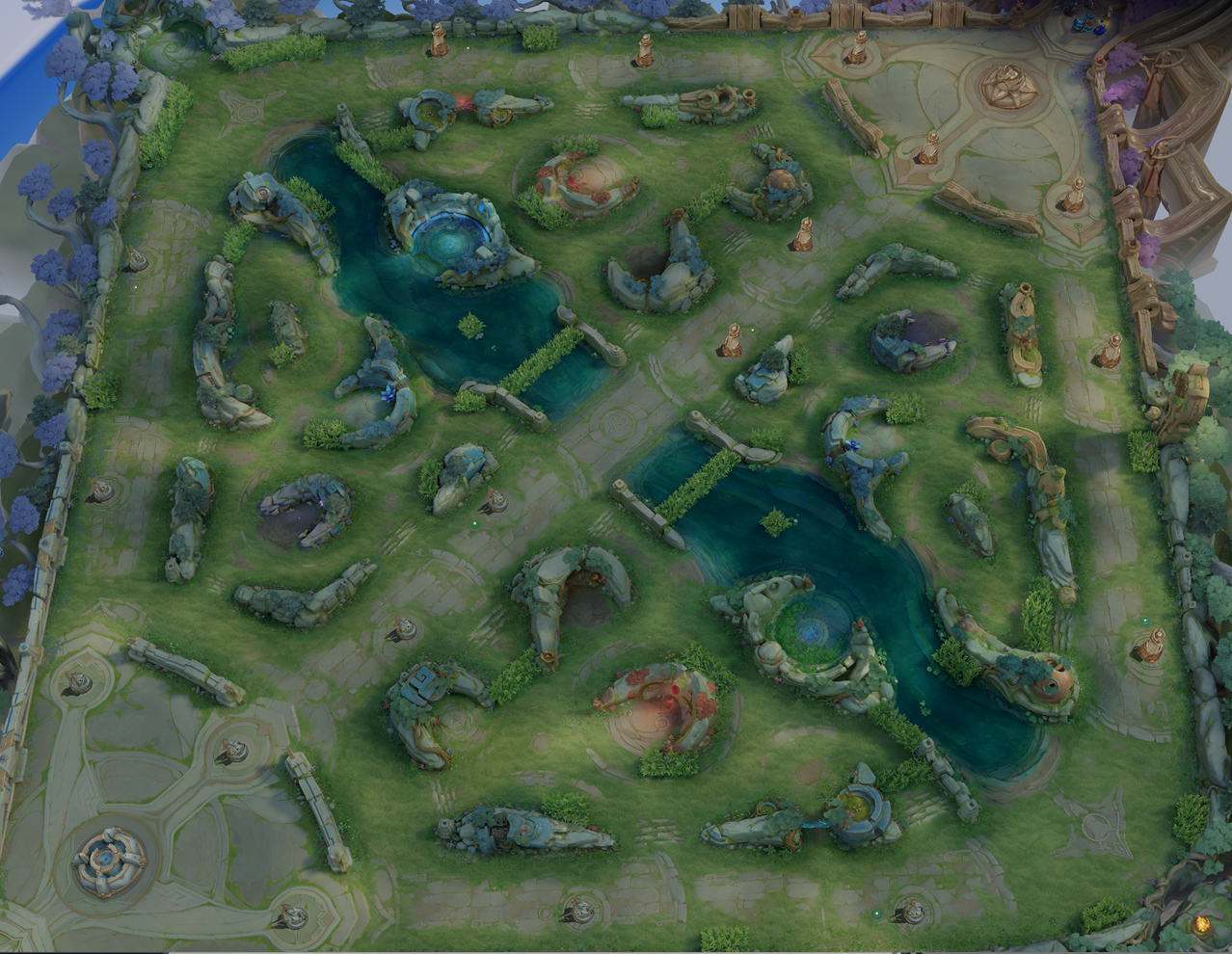}
\caption{The HoK game map.}
\label{fig:hok_map}
\end{figure}

\subsection{Dynamic Event Taxonomy and Trigger Protocols}
\label{app:dyn_events}
\subsubsection{The design rationale of dynamic events}
Real-world dynamic events form an open-ended combinatorial perturbation space: enemy support can arrive from any direction with any composition; held items can drop at any frame; an ally can disconnect or be teleported mid-chain. Enumerating this space exhaustively is impossible. Instead, for evaluation we define a small set of \emph{dynamic event types} that provides broad coverage of the main failure modes. Each type is triggered by a seeded script in Overcooked so that the same perturbation can be replayed across methods and seeds.
\begin{figure*}[htbp]
    \centering
    \includegraphics[width=0.95\linewidth]{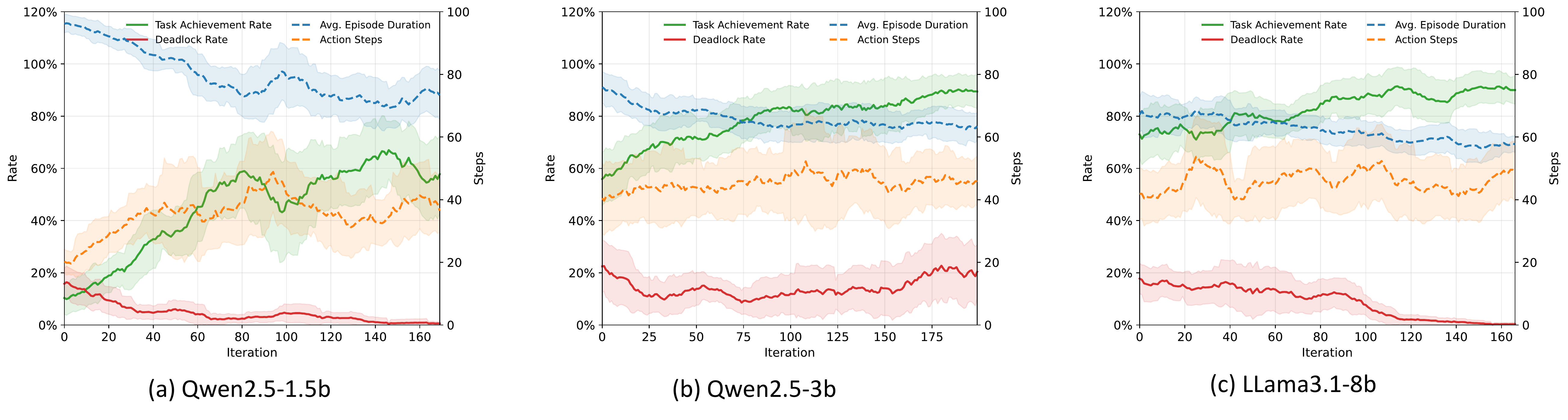}
    \caption{In-game performance variation over the course of RL training.}
    \label{fig:game}
\end{figure*}
\subsubsection{Overcooked dynamic event injection}
\label{app:oc_events}

Table~\ref{tab:oc_events_full} summarizes the three Overcooked dynamic event types and the action-chain field each one stresses.
Overcooked events reuse the injection hooks exposed by the Collab-Overcooked platform. All three events are triggered from a seeded RNG so that runs across methods and seeds see the exact same sequence.

\begin{table}[htbp]
\centering
\footnotesize
\setlength{\tabcolsep}{3pt}
\resizebox{\columnwidth}{!}{%
\begin{tabular}{lll}
\toprule
\textbf{ID} & \textbf{Name}  & \textbf{Example} \\
\midrule
O1 & Held-Item Drop          & Held onion drops to floor \\
O2 & Pot-Content Scatter  & 1 of 3 onions scattered; timer resets \\
O3 & Agent Displacement   & Agent teleported to a far counter \\
\bottomrule
\end{tabular}%
}
\caption{Overcooked dynamic events.}
\label{tab:oc_events_full}
\end{table}

\begin{table}[htbp]
\centering
\footnotesize
\scalebox{1}{
\begin{tabular}{lc}
\toprule
\textbf{Hyperparameter} & \textbf{Value} \\
\midrule
\multicolumn{2}{l}{\textit{Model}} \\
\quad Flash attention & FA2 \\
\quad Precision & bfloat16 \\
\midrule
\multicolumn{2}{l}{\textit{LoRA}} \\
\quad Rank & 32 \\
\quad Alpha & 64 \\
\quad Dropout & 0.05 \\
\quad Target modules & All linear layers \\
\midrule
\multicolumn{2}{l}{\textit{Data}} \\
\quad Template & Qwen chat format \\
\quad Max sequence length & 4,096 tokens \\
\quad Validation split & 5\% \\
\midrule
\multicolumn{2}{l}{\textit{Optimization}} \\
\quad Per-device batch size & 1 \\
\quad Gradient accumulation & 8 \\
\quad Learning rate & $2 \times 10^{-5}$ \\
\quad LR schedule & Cosine \\
\quad Warmup ratio & 0.1 \\
\quad Epochs & 4 \\
\quad Save interval & 200 steps \\
\quad Eval interval & 200 steps \\
\bottomrule
\end{tabular}
}
\caption{SFT training hyperparameters (LLaMA-Factory + LoRA).}
\label{tab:sft_hyperparams}
\end{table}
\subsubsection{HoK dynamic event injection} In HoK we do \emph{not} inject any scripted events. The environment is inherently non-stationary: uncontrolled allied and adversarial bots act on their own policies, so perturbations to an action chain (enemy reinforcements rotating in, an objective disengaging, a teammate dying) arise naturally during play. HoK therefore serves as a complexity stress-test in which dynamic events are endogenous, and only Overcooked uses the injected event protocol below.

\begin{table}[htbp]
\centering
\footnotesize
\scalebox{1}{
\begin{tabular}{lc}
\toprule
\textbf{Hyperparameter} & \textbf{Value} \\
\midrule
\multicolumn{2}{l}{\textit{Model}} \\
\quad Base model & Qwen2.5-7B-Instruct \\
\quad Max sequence length & 4,096 tokens \\
\quad Precision & bfloat16 \\
\midrule
\multicolumn{2}{l}{\textit{LoRA}} \\
\quad Rank & 32 \\
\quad Alpha & 64 \\
\quad Target modules & q, k, v, o, gate, up, down proj \\
\midrule
\multicolumn{2}{l}{\textit{GRPO}} \\
\quad Prompts per batch & 10 \\
\quad Responses per prompt & 10 \\
\quad Min group size & 2 \\
\quad Clipping threshold $\epsilon$ & 0.2 \\
\quad KL coefficient $\beta$ & 0.02 \\
\quad Entropy coefficient & 0.001 \\
\midrule
\multicolumn{2}{l}{\textit{Optimization}} \\
\quad Learning rate & $5 \times 10^{-5}$ \\
\quad Warmup ratio & 0.03 \\
\quad Per-device batch size & 2 \\
\quad Gradient accumulation & 4 \\
\quad Max grad norm & 1.0 \\
\quad Epochs & 1 \\
\quad Seed & 42 \\
\bottomrule
\end{tabular}
}
\caption{RL (GRPO) training hyperparameters.}
\label{tab:hyperparams}
\end{table}

\begin{figure*}[htbp]
    \centering
    \includegraphics[width=0.95\linewidth]{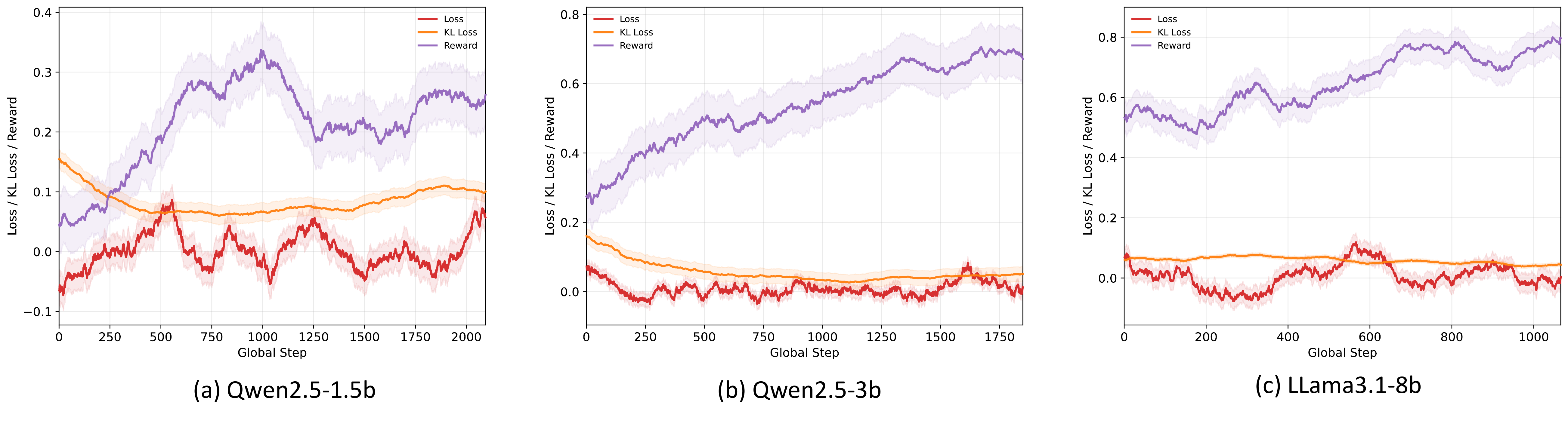}
    \caption{Training curves (reward and loss) during RL optimization.}
    \label{fig:train_rl}
\end{figure*}
\subsection{SFT Training Hyperparameters}
\label{app:sft_hyperparams}

Table~\ref{tab:sft_hyperparams} lists the hyperparameters for the supervised fine-tuning stage. We use LLaMA-Factory with LoRA adaptation on Qwen2.5-7B-Instruct.
\begin{figure*}[!htbp]
    \centering
    \includegraphics[width=0.9\linewidth]{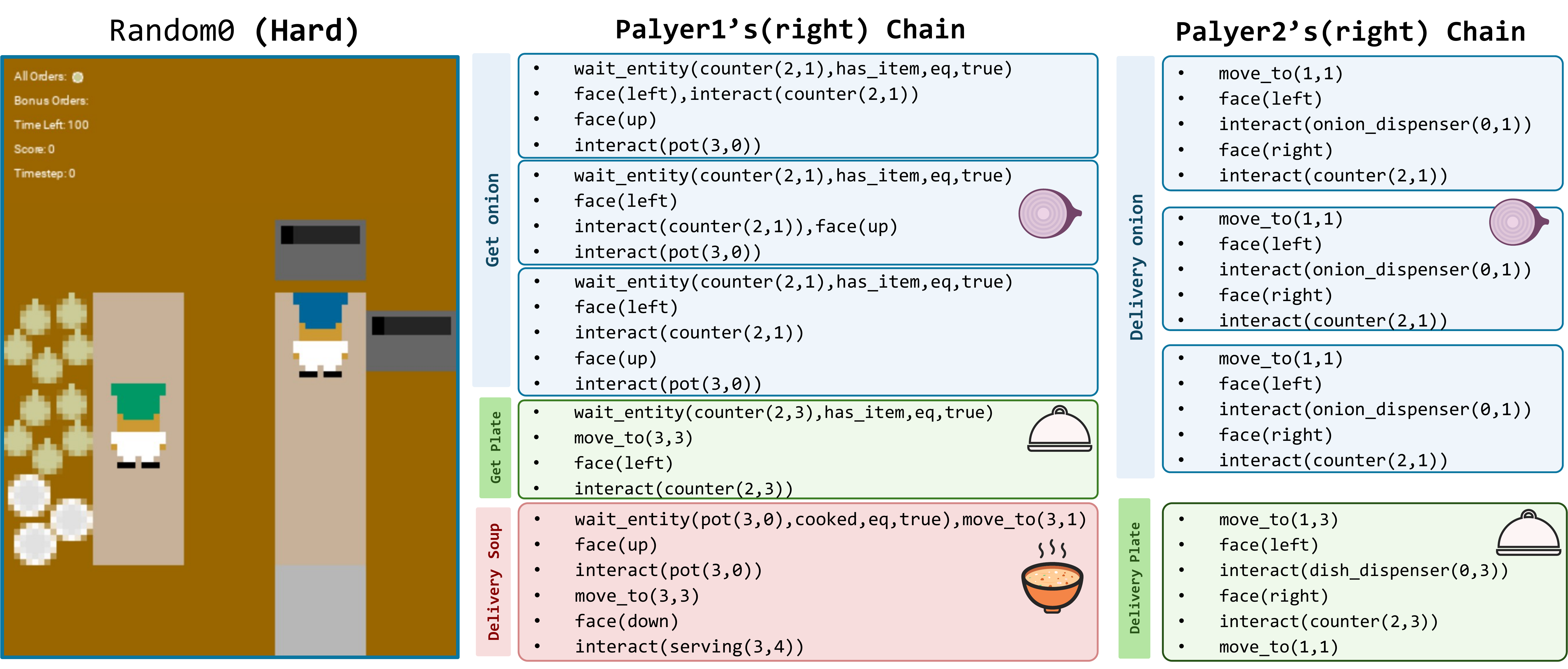}
    \caption{Overcooked case study on the \emph{Coordination} layout. When the active chains create mutually dependent \texttt{Wait\_agents} predicates, the incremental reachability test in Section~\ref{app:deadlock} detects the resulting wait cycle. The re-invoked Plan Module emits a revised chain with an explicit \texttt{Wait\_agents} step that serializes tile access, allowing both agents to complete delivery without collision.}
    \label{fig:case_overcooked}
\end{figure*}
\subsection{RL Training Hyperparameters}
\label{app:hyperparams}
Table~\ref{tab:hyperparams} lists all hyperparameters for the GRPO-based RL training.
\subsection{PSD Architecture}
\begin{table}[htbp]
  \centering
  \caption{PSD Architecture}
\resizebox{\linewidth}{!}{
\begin{tabular}{ccccccc}
    \toprule
    Arch. & Loss  & Param & F1(opt) & Acc(opt) & Precision & Recall \\
    \midrule
    \multirow{2}[1]{*}{Small} & BCE   & 68,737 & 0.8062 & 0.8363 & 0.6799 & 0.9902 \\
          & Focal & 68,737 & 0.8189 & 0.8491 & 0.697 & 0.9924 \\
    \multirow{2}[0]{*}{Medium} & BCE   & 261,377 & 0.8284 & 0.8588 & 0.7115 & 0.9913 \\
          & Focal & 261,377 & 0.8204 & 0.8509 & 0.7003 & 0.9902 \\
    \multirow{2}[1]{*}{Wide} & BCE   & 745,473 & 0.8169 & 0.8472 & 0.6947 & 0.9913 \\
          & Focal & 745,473 & 0.8382 & 0.8685 & 0.7266 & 0.9902 \\
    \bottomrule
    \end{tabular}%
    }
  \label{tab:psd}%
\end{table}%
To determine the optimal architecture for the Plan Staleness Detector (PSD), we conducted an architecture search across different model scales and loss functions. As shown in Table~\ref{tab:psd}, we evaluated three architectures—Small (68K parameters), Medium (261K parameters), and Wide (745K parameters)—with both Binary Cross-Entropy (BCE) and Focal loss. The results indicate that performance differences across architectures are relatively modest, with F1 scores ranging from 0.8062 to 0.8382 and accuracy from 0.8363 to 0.8685. Notably, all configurations achieve consistently high recall (above 0.99), which is critical for detecting plan staleness to avoid coordination failures. The Wide architecture with Focal loss achieves the best overall performance (F1: 0.8382, Acc: 0.8685), offering a favorable trade-off between precision (0.7266) and recall (0.9902). Based on these findings, we adopt the Wide+Focal configuration for subsequent experiments, as the marginal parameter increase is justified by its superior detection capability.

\begin{figure}[!htbp]
    \centering
    \includegraphics[width=1\linewidth]{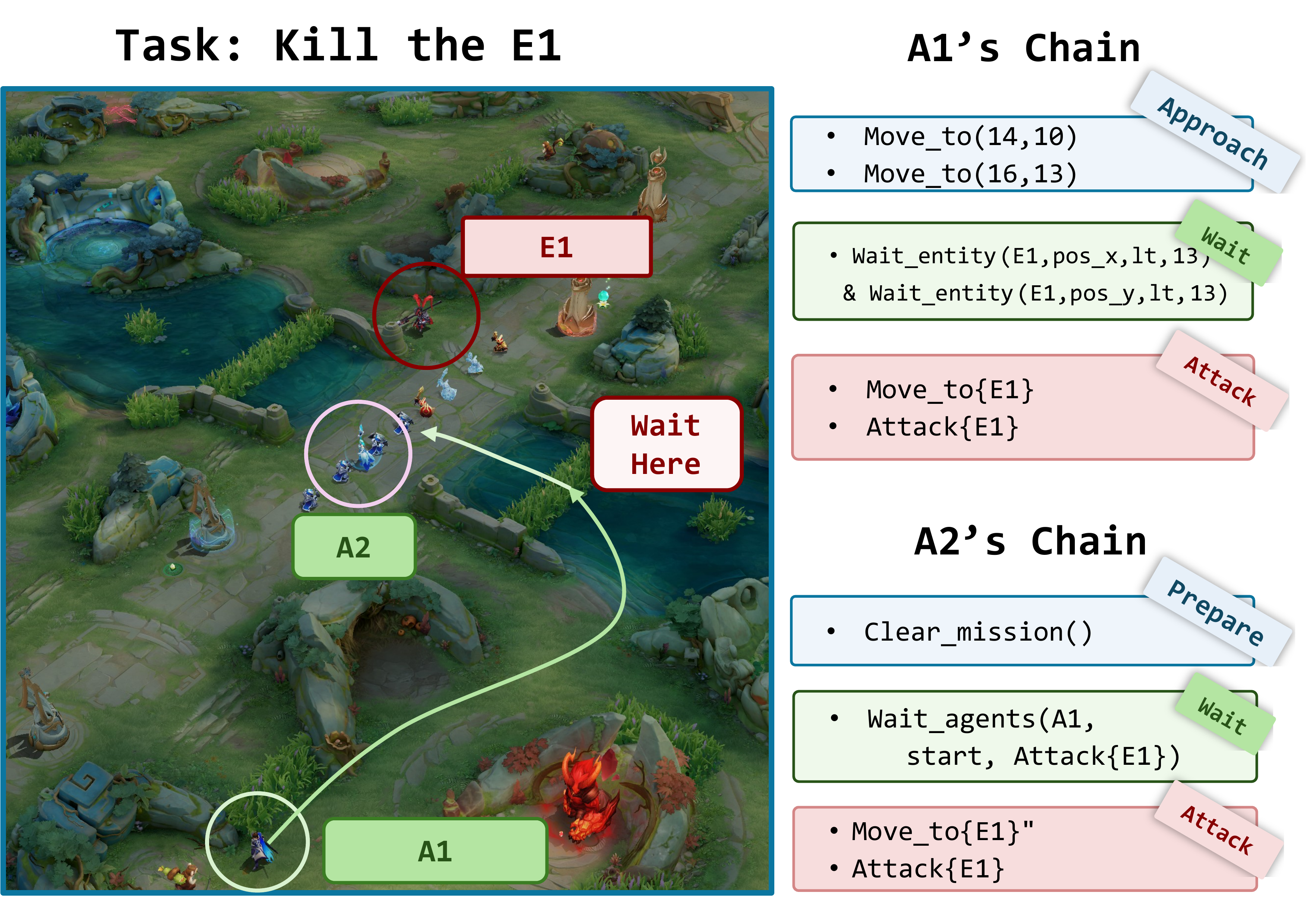}
    \caption{Honor of Kings case study: the PSD detects an unexpected health drop on the support hero caused by an enemy rotation invalidating the original engagement-window assumption.}
    \label{fig:case_hok}
\end{figure}

\section{Additional Experimental Analysis}

\subsection{Training Dynamics}
\label{app:training_dynamics}

We visualize the RL training process from two complementary perspectives: in-game task performance (Figure~\ref{fig:game}) and optimisation-level statistics (Figure~\ref{fig:train_rl}).

\paragraph{In-game performance.} Figure~\ref{fig:game} tracks task completion and deadlock-rate trajectories over GRPO iterations. The curves show improving task completion together with a declining rate of cyclic wait failures. They are descriptive training diagnostics rather than an estimate of generalization to unseen event distributions.

\paragraph{Optimization curves.} Figure~\ref{fig:train_rl} reports the GRPO reward and policy-loss trajectories. The reward trend is interpreted jointly with the task-level evaluation tables: it is evidence of training progress, but it is not by itself evidence that any single reward component or module causes the final performance gain.

\subsection{Case Studies}
\label{app:cases}
This section expands the two representative episodes summarized in the main paper's Case Studies section with the full step-by-step trace of each correction event.

\subsubsection{Overcooked: Resolving a Spatial Conflict}

Figure~\ref{fig:case_overcooked} traces an episode on the Coordination layout in which the initial plan assigns both agents an overlapping tile as an intermediate waypoint. If execution turns this conflict into mutually dependent \textbf{\emph{Wait\_agents}} predicates, the incremental reachability check in Section~\ref{app:deadlock} detects that adding the new wait edge closes a cycle and raises a graph-deadlock signal. The Plan Module is then reinvoked and produces an explicit \textbf{\emph{Wait\_agents}} step that serializes access to the shared tile, after which the agents complete their pickup/delivery sequence.

\subsubsection{Honor of Kings: PSD-Triggered Reactive Correction}

Figure~\ref{fig:case_hok} traces a team-gank episode in which the original plan assumes a safe engagement window for a flanking maneuver on the enemy jungler. Partway through execution, an off-screen enemy hero rotates into the fight and the support hero's health changes unexpectedly relative to the state on which the chain was conditioned. Because the affected hero is referenced by a remaining chain step, the PSD state pathway incorporates this entity-level change; once $P_{\mathrm{replan}}$ exceeds $\theta_{\mathrm{PSD}}=0.7$, the Plan Module is reinvoked on the updated state. The resulting chain redirects the support hero to disengage and provide healing rather than continuing the original flanking route, while the remaining agents adjust their engagement timing. This case illustrates selective correction under endogenous HoK dynamics; aggregate latency results are reported in the main paper.

\end{document}